%% file: main.tex
\documentclass[10pt,twocolumn,letterpaper]{article}

\usepackage[pagenumbers]{cvpr} 

\input{preamble}

\definecolor{cvprblue}{rgb}{0.21,0.49,0.74}
\usepackage[pagebackref,breaklinks,colorlinks,citecolor=cvprblue]{hyperref}

\usepackage{hyperref}
\usepackage{url}

\usepackage{graphicx}
\usepackage{amsmath}
\usepackage{amssymb}
\usepackage{booktabs}

\usepackage{colortbl}
\usepackage{xcolor}
\usepackage{array}

\usepackage{adjustbox}
\usepackage{times}
\usepackage{epsfig}
\usepackage{graphicx}
\usepackage{amsmath}
\usepackage{amssymb}
\usepackage{booktabs}
\usepackage{colortbl}
\usepackage{tabularx}
\usepackage{tabu}
\usepackage{xcolor}
\usepackage{multirow}
\usepackage{graphicx} 
\usepackage{multicol}
\usepackage{float}
\usepackage{enumitem}
\usepackage{placeins}
\usepackage{tabularx}

\usepackage[accsupp]{axessibility} 

\newcolumntype{C}{>{\centering\arraybackslash}X}

\newcommand{\ours}{TADP}

\definecolor{CBLightCyan}{HTML}{bae1ff}
\definecolor{CBLightYellow}{HTML}{ffffba}
\definecolor{CBLightGreen}{HTML}{ddf2d1}
\definecolor{LightCyan}{rgb}{0.88,1,1}

\newcommand{\model}{\epsilon_\theta}
\newcommand{\conditioner}{\tau_\theta}
\newcommand{\expec}{\mathbb{E}}

\newcommand{\encoder}{\mathcal{E}}
\newcommand{\decoder}{\mathcal{D}}
\newcommand{\capmod}{\mathcal{M}}
\newcommand{\captioner}{\mathcal{G}}
\newcommand{\capt}{\mathcal{C}}
\newcommand{\cnset}{\mathcal{B}}
\newcommand{\tdinf}{\mathcal{P}}
\newcommand{\lsimplelcm}{L_{LDM}}

\def\paperID{8551} 
\def\confName{CVPR}
\def\confYear{2024}

\title{Text-image Alignment for Diffusion-based Perception}

\author{
Neehar Kondapaneni$^{1}$\thanks{Equal contribution.}
~~
Markus Marks$^{1*}$
~~
Manuel Knott$^{1,2*}$
\\
Rogerio Guimaraes$^{1}$
~~
Pietro Perona$^{1}$
\\\\
$^{1}$California Institute of Technology
\\
$^{2}$ETH Zurich, Swiss Data Science Center, Empa
}

\newcommand{\beginsupplement}{%
        \setcounter{table}{0}
        \renewcommand{\thetable}{S\arabic{table}}%
        \setcounter{figure}{0}
        \renewcommand{\thefigure}{S\arabic{figure}}%
        \setcounter{section}{0}
        \renewcommand{\thesection}{\Alph{section}}%
     }

\begin{document}
\maketitle

\begin{abstract}
Diffusion models are generative models with impressive text-to-image synthesis capabilities and have spurred a new wave of creative methods for classical machine learning tasks. However, the best way to harness the perceptual knowledge of these generative models for visual tasks is still an open question.
Specifically, it is unclear how to use the prompting interface when applying diffusion backbones to vision tasks. We find that automatically generated captions can improve text-image alignment and significantly enhance a model's cross-attention maps, leading to better perceptual performance. Our approach improves upon the current state-of-the-art (SOTA) in diffusion-based semantic segmentation on ADE20K and the current overall SOTA for depth estimation on NYUv2. 
Furthermore, our method generalizes to the cross-domain setting. We use model personalization and caption modifications to align our model to the target domain and find improvements over unaligned baselines. Our cross-domain object detection model, trained on Pascal VOC, achieves SOTA results on Watercolor2K. Our cross-domain segmentation method, trained on Cityscapes, achieves SOTA results on Dark Zurich-val and Nighttime Driving. Project page: \href{https://www.vision.caltech.edu/tadp/}{\texttt{vision.caltech.edu/TADP/}}
Code page: \href{https://github.com/damaggu/TADP}{\texttt{github.com/damaggu/TADP}}

\end{abstract}

\vspace{-10pt}
\section{Introduction}
\label{sec:intro}

\begin{figure}
    \centering
    \vspace{\baselineskip}
    \includegraphics[trim={0.2cm 1.9cm 4.2cm 0.3cm},clip, width=\linewidth]{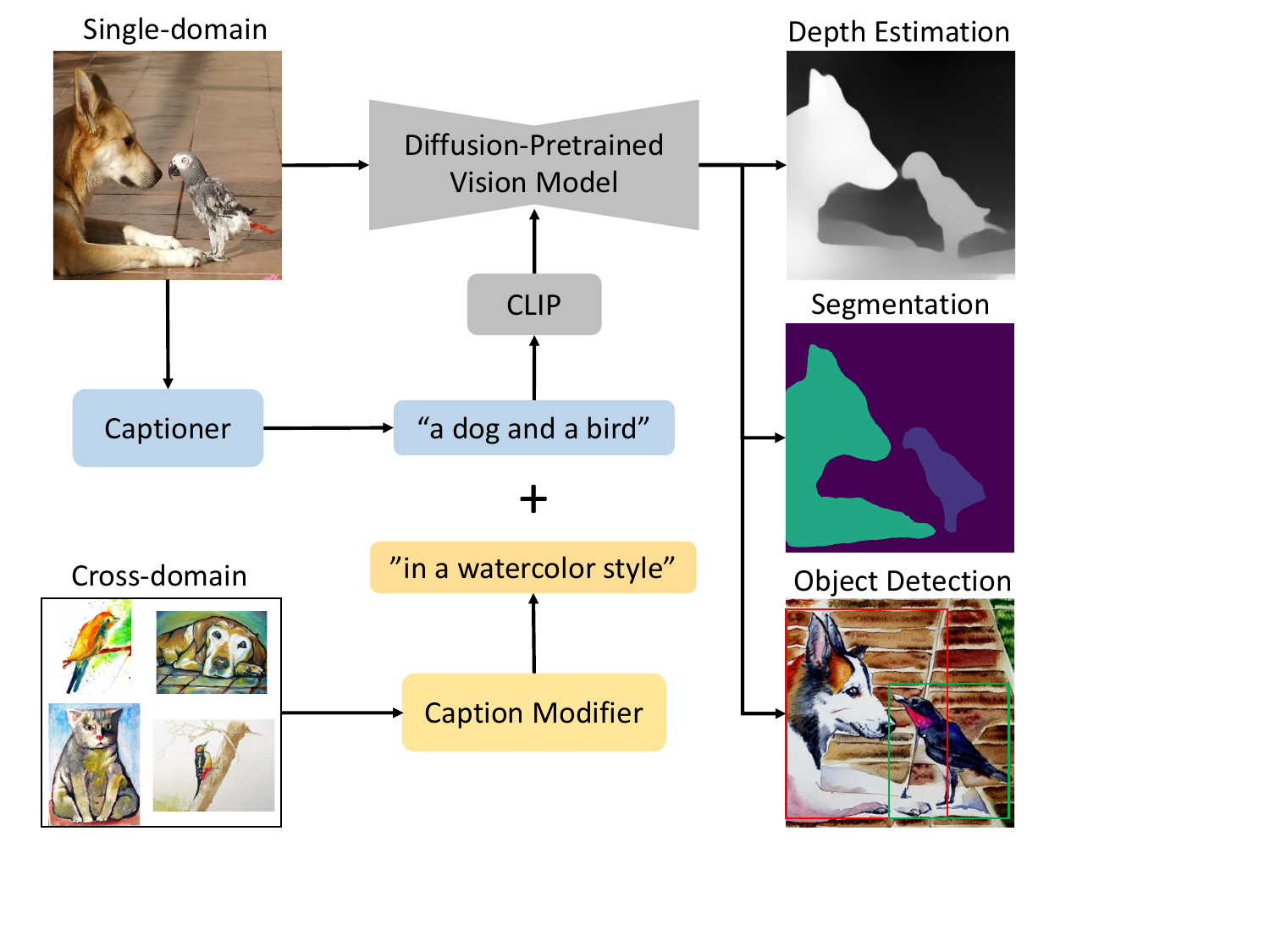}
    \caption{\textbf{\textbf{T}ext-\textbf{A}ligned \textbf{D}iffusion \textbf{P}erception (TADP).} In TADP, image captions align the text prompts and images passed to diffusion-based vision models. In cross-domain tasks, target domain information is incorporated into the prompt to boost performance.}
    \label{fig:overview}
\end{figure}

Diffusion models have set the state-of-the-art (SOTA) for image generation \citep{yu_scaling_2022, ramesh_hierarchical_2022, saharia_photorealistic_2022, rombach_high-resolution_2022}. Recently, a few works have shown diffusion pre-trained backbones have a strong prior for scene understanding that allows them to perform well in advanced discriminative vision tasks, such as semantic segmentation \citep{gong_prompting_2023, zhao_unleashing_2023}, monocular depth estimation \citep{zhao_unleashing_2023}, and keypoint estimation \citep{luo2023diffusion, tang2023emergent}. We refer to these works as diffusion-based perception methods. Unlike contrastive vision language models (e.g., CLIP) \citep{radford_learning_2021, jia_scaling_2021, li_supervision_2022}, generative models have a causal relationship with text, in which text guides image generation. In latent diffusion models, text prompts control the denoising U-Net \citep{navab_u-net_2015}, moving the image latent in a semantically meaningful direction \citep{brack_sega_2023}. 

We explore this relationship and find that text-image alignment significantly improves the performance of diffusion-based perception. We then investigate text-target domain alignment in cross-domain vision tasks, finding that aligning to the target domain while training on the source domain can improve a model's target domain performance (\cref{fig:overview}).

We first study prompting for diffusion-based perceptual models and find that increasing text-image alignment improves semantic segmentation and depth estimation performance. We find that unaligned text prompts can introduce semantic shifts to the feature maps of the diffusion model \citep{brack_sega_2023} and that these shifts can make it more difficult for the task-specific head to solve the target task. Specifically, we ask whether unaligned text prompts, such as averaging class-specific sentence embeddings (\cite{radford_learning_2021, zhao_unleashing_2023}), hinder performance by interfering with feature maps through the cross-attention mechanism. Through ablation experiments on Pascal VOC2012 segmentation \citep{everingham_pascal_2012} and ADE20K \citep{zhou_scene_2017}, we find that off-target and missing class names degrade image segmentation quality. We show automated image captioning \citep{li_blip-2_2023} achieves sufficient text-image alignment for perception. Our approach (along with latent representation scaling, see \cref{latent_scaling}) improves performance for semantic segmentation on Pascal and ADE20k by 4.0 mIoU and 1.7 mIoU, respectively, and depth estimation on NYUv2 \citep{hutchison_indoor_2012} by 0.2 RMSE (+8\% relative) setting the new SOTA.

Next, we focus on cross-domain adaptation: can appropriate image captioning help visual perception when the model is trained in one domain and tested on a different domain? Training models on the source domain with the appropriate prompting strategy leads to excellent unsupervised cross-domain performance on several benchmarks. We evaluate our cross-domain method on Pascal VOC \citep{pascal-voc-2007, everingham_pascal_2012} to Watercolor2k (W2K) and Comic2k (C2K) \citep{inoue_cross-domain_2018} for object detection and Cityscapes (CS) \citep{cordts_cityscapes_2016} to Dark Zurich (DZ) \citep{sakaridis_guided_2019} and Nighttime (ND) Driving \citep{dai_dark_2018} for semantic segmentation. We explore varying degrees of text-target domain alignment and find that improved alignment results in better performance. We also demonstrate using two diffusion personalization methods, Textual Inversion \citep{gal_image_2022} and DreamBooth \citep{ruiz_dreambooth_2022}, for better target domain alignment and performance. We find that diffusion pre-training is sufficient to achieve SOTA (+5.8 mIoU on CS$\to$DZ, +4.0 mIoU on CS$\to$ND, +0.7 mIoU on VOC$\to$W2k) or near SOTA results on all cross-domain datasets with no text-target domain alignment, and including our best text-target domain alignment method further improves +1.4 AP on Watercolor2k, +2.1 AP on Comic2k, and +3.3 mIoU on Nighttime Driving.
\\Overall, our contributions are as follows:
\begin{itemize}[itemsep=-0.25ex,leftmargin=0.3in]
    \item We propose a new method using automated caption generation that significantly improves performance on several diffusion-based vision tasks through increased text-image alignment.
    \item We systematically study how prompting affects diffusion-based vision performance, elucidating the impact of class presence, grammar in the prompt, and previously used average embeddings.
    \item We demonstrate that diffusion-based perception effectively generalizes across domains, with text-target domain alignment improving performance, which can be further boosted by model personalization. 
\end{itemize}

\section{Related Work}

\subsection{Diffusion models for single-domain vision tasks}

Diffusion models are trained to reverse a step-wise forward noising process. Once trained, they can generate highly realistic images from pure noise \citep{rombach_high-resolution_2022, ramesh_hierarchical_2022, yu_scaling_2022, saharia_photorealistic_2022}. To control image generation, diffusion models are trained with text prompts/captions that guide the diffusion process. These prompts are passed through a text encoder to generate text embeddings that are incorporated into the reverse diffusion process via cross-attention layers.

Recently, some works have explored using diffusion models for discriminative vision tasks. This can be done by either utilizing the diffusion model as a backbone for the task \citep{zhao_unleashing_2023, gong_prompting_2023, tang2023emergent, luo2023diffusion} or through fine-tuning the diffusion model for a specific task and then using it to generate synthetic data for a downstream model \citep{wu_diffumask_2023, benigmim_one-shot_2023}. We use the diffusion model as a backbone for downstream vision tasks.

VPD \citep{zhao_unleashing_2023} encodes images into latent representations and passes them through one step of the Stable Diffusion model. The cross-attention maps, multi-scale features, and output latent code are concatenated and passed to a task-specific head. Text prompts influence all these maps through the cross-attention mechanism, which guides the reverse diffusion process. The cross-attention maps are incorporated into the multi-scale feature maps and the output latent representation. The text guides the diffusion process and can accordingly shift the latent representation in semantic directions \citep{brack_sega_2023, gal_image_2022, hertz_prompt--prompt_2022, balaji_ediff-i_2022}. The details of how VPD uses the prompting interface are described in \cref{methods_vpd}. In short, VPD uses \textit{unaligned} text prompts. In our work, we show how aligning the text to the image by using a captioner can significantly improve semantic segmentation and depth estimation performance.

\subsection{Image captioning}
CLIP \citep{radford_learning_2021} introduced a novel learning paradigm to align images with their captions. Shortly after, the LAION-5B dataset \citep{schuhmann_laion-5b_2022} was released with 5B image-text pairs; this dataset was used to train Stable Diffusion. We hypothesize that text-image alignment is important for diffusion-pretrained vision models. However, images used in advanced vision tasks (like segmentation and depth estimation) are not naturally paired with text captions. To obtain image-aligned captions, we use BLIP-2 \citep{li_blip-2_2023}, a model that inverts the CLIP latent space to generate captions for novel images. 

\subsection{Diffusion models for cross-domain vision tasks}
A few works explore the cross-domain setting with diffusion models \citep{benigmim_one-shot_2023, gong_prompting_2023}. \citet{benigmim_one-shot_2023} use a diffusion model to generate data for a downstream unsupervised domain adaptation (UDA) architecture.
In \cite{gong_prompting_2023}, the diffusion backbone is frozen, and the segmentation head is trained with a consistency loss with category and scene prompts guiding the latent code towards target cross-domains. Similar to VPD, the category prompts consist of token embeddings for all classes present in the dataset, irrespective of their presence in any specific image. The consistency loss forces the model to predict the same output mask for all the different scene prompts, helping the segmentation head become invariant to the scene type. Instead of using a consistency loss, we train the diffusion model backbone and task head on the source domain data with and without incorporating the style of the target domain in the caption. We find that better alignment with the target domain (i.e., target domain information included in the prompt) results in better cross-domain performance.

\subsection{Cross-domain object detection}

Cross-domain object detection can be divided into multiple subcategories, depending on what data~/~labels are at train~/~test time available. Unsupervised domain adaptation objection detection (UDAOD) tries to improve detection performance by training on unlabeled target domain data with approaches such as self-training \citep{deng2021unbiased, topal2022domain}, adversarial distribution alignment \citep{zhao2020adaptive} or generating pseudo labels for self-training \citep{jiang2021decoupled}. Cross-domain weakly supervised object detection (CDWSOD) assumes the availability of image-level annotations at training time and utilizes pseudo labeling  \citep{ouyang2021pseudo, inoue_cross-domain_2018}, alignment \citep{xu2022h2fa} or correspondence mining \citep{hou2021informative}. 
Recently, \citep{vidit_clip_2023} used CLIP \citep{radford_learning_2021} for Single Domain Generalization, which aims to generalize from a single domain to multiple unseen target domains. Our text-based method defines a new category of cross-domain object detection that tries to adapt from a single source to an unseen target domain by only having the broad semantic context of the target domain (e.g., foggy/night/comic/watercolor) as text input to our method. When we incorporate model personalization, our method can be considered a UDAOD method since we train a token based on unlabeled images from the target domain.

\section{Methods}
\label{methods}

\begin{figure*}[t!]
    \centering
    \includegraphics[width=\textwidth]{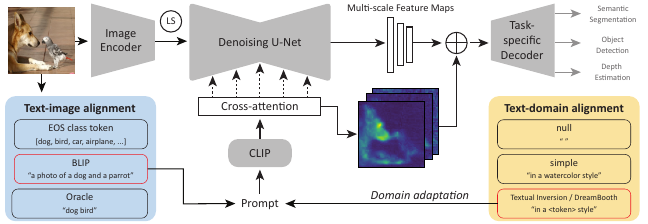}
    \caption{\textbf{Overview of TADP.} We test several prompting strategies and evaluate their impact on downstream vision task performance. Our method concatenates the cross-attention and multi-scale feature maps before passing them to the vision-specific decoder. In the blue box, we show three single-domain captioning strategies with differing levels of text-image alignment. We propose using BLIP \citep{li_blip-2_2023} captioning to improve image-text alignment. We extend our analysis to the cross-domain setting (yellow box), exploring whether aligning the source domain text captions to the target domain may impact model performance by appending caption modifiers to image captions generated in the source domain and find model personalization modifiers (Textual Inversion/Dreambooth) work best.}
    \label{fig:coremethods}
    \vspace{-10pt}
\end{figure*}

\label{methods_sd}
\textbf{Stable Diffusion \citep{rombach_high-resolution_2022}.} The text-to-image Stable Diffusion model is composed of four networks: an encoder $\encoder$, a conditional denoising autoencoder (a U-Net in Stable Diffusion) $\model$, a language encoder $\conditioner$ (the CLIP text encoder in Stable Diffusion), and a decoder $\decoder$. $\encoder$ and $\decoder$ are trained before $\model$, such that $\decoder(\encoder(x)) = \Tilde{x} \approx x$. Training $\model$ is composed of a pre-defined forward process and a learned reverse process. The reverse process is learned using LAION-400M \citep{schuhmann_laion-400m_2021}, a dataset of 400 million images ($x \in X$) and captions ($y \in Y$). In the forward process, an image $x$ is encoded into a latent $z_0 = \encoder(x)$, and $t$ steps of a forward noise process are executed to generate a noised latent $z_t$. Then, to learn the reverse process, the latent $z_t$ is passed to the denoising autoencoder $\model$, along with the time-step $t$ and the image caption's representation $\capt=\conditioner(y)$. $\conditioner$ adds information about $y$ to $\model$ using a cross-attention mechanism, in which the query is derived from the image, and the key and value are transformations of the caption representation. The model $\model$ is trained to predict the noise added to the latent in step $t$ of the forward process: 
\begin{equation}
\lsimplelcm := \expec_{\encoder(x), y, \epsilon \sim \mathcal{N}(0, 1), t }\Big[ \Vert \epsilon - \model(z_{t},t, \conditioner(y)) \Vert_{2}^{2}\Big] \, ,
\label{eq:cond_loss}
\end{equation}
where $t \in \{0,...,T\}$. During generation, a pure noise latent $z_T$ and a user-specified prompt are passed through the denoising autoencoder $\model$ for $T$ steps and decoded $\decoder(z_0)$ to generate an image guided by the text prompt.

\label{methods_vpd}
\textbf{Diffusion for Feature Extraction.} Diffusion backbones have been used for downstream vision tasks in several recent works~\cite{zhao_unleashing_2023, gong_prompting_2023, luo2023diffusion, tang2023emergent}. Due to its public availability and performance in perception tasks, we use a modified version (see \cref{latent_scaling}) of the feature extraction method in VPD. An image latent $z_0 = \encoder(x)$ and a conditioning $\mathcal{C}$ are passed through the last step of the denoising process $\model(z_0, 0, \mathcal{C})$. The cross-attention maps $A$ and the multi-scale feature maps $F$ of the U-Net are concatenated $V = A \oplus F$ and passed to a task-specific head $H$ to generate a prediction $\hat{p} = H(V)$. The backbone $\model$ and head $H$ are trained with a task-specific loss $\mathcal{L}_{H}(\hat{p}, p)$. 

\label{methods_prompts}
\textbf{Average EOS Tokens.}
To generate $\mathcal{C}$, previous methods~\cite{zhao_unleashing_2023, gong_prompting_2023} rely on a method from CLIP~\cite{radford_learning_2021} to use averaged text embeddings as representations for the classes in a dataset. A list of 80 sentence templates for each class of interest (such as ``a \textless adjective\textgreater\ photo of a \textless class name\textgreater'') are passed through the CLIP text encoder. We use $\cnset$ to denote the set of class names in a dataset. For a specific class ($b \in \cnset$), the CLIP text encoder returns an $80 \times N\times D$ tensor, where N is the maximum number of tokens over all the templates, and D is 768 (the dimension of each token embedding). Shorter sentences are padded with EOS tokens to fill out the maximum number of tokens. The first EOS token from each sentence template is averaged and used as the representative embedding for the class such that $\capt \in \mathcal{R}^{|\cnset| \times 768}$. This method is used in~\cite{zhao_unleashing_2023, gong_prompting_2023}, we denote it as $\mathcal{C}_{avg}$ and use it as a baseline.
For semantic segmentation, all of the class embeddings, irrespective of presence in the image, are passed to the cross-attention layers. Only the class embedding of the room type is passed to the cross-attention layers for depth estimation.

\subsection{Text-Aligned Diffusion Perception (TADP)}
\label{methods_tadp}
Our work proposes a novel method for prompting diffusion-pretrained perception models. Specifically, we explore different prompting methods $\captioner$ to generate $\mathcal{C}$. In the single-domain setting, we show the effectiveness of a method that uses BLIP-2 ~\citep{li_blip-2_2023}, an image captioning algorithm, to generate a caption as the conditioning for the model: $\captioner(x) = \Tilde{y} \rightarrow \capt$. We then extend our method to the cross-domain setting by incorporating target domain information to $\mathcal{C} = \mathcal{C} + \capmod(\tdinf)_s$, where $\capmod$ is a caption modifier that takes target domain information $\tdinf$ as input and outputs a caption modification $\capmod(\tdinf)_s$ and a model modification $\capmod(\tdinf)_{\model}$. In \cref{results}, we analyze the text-image interface of the diffusion model by varying the captioner $\captioner$ and caption modifier $\capmod$ in a systematic manner for three different vision tasks: semantic segmentation, object detection, and monocular depth estimation. Our method and experiments are presented in \cref{fig:coremethods}. Following~\cite{zhao_unleashing_2023}, we train our ADE20k segmentation and NYUv2 depth estimation models with fast and regular schedules. On ADE20k, we train using 4k steps (fast), 8k steps (fast), and 80k steps (normal). For NYUv2 depth, we train on a 1-epoch (fast) schedule and a 25-epoch (normal) schedule. For implementation details, refer to Appendix~\ref{appendix:implementation_details}.

\section{Results}
\label{results}

\input{tables/seg_pascal}
\subsection{Latent scaling}
\label{latent_scaling}

Before exploring image-text alignment, we apply latent scaling to encoded images (Appendix~G of \citet{rombach_high-resolution_2022}). This normalizes the image latents to have a standard normal distribution. The scaling factor is fixed at $0.18215$. We find that latent scaling improves performance using $\mathcal{C}_{avg}$ for segmentation and depth estimation (\cref{fig:bar_plots}). Specifically, latent scaling improves $\sim$0.8\%~mIoU on Pascal, $\sim$0.3\% mIoU on ADE20K, and a relative $\sim$5.5\% RMSE on NYUv2 Depth (\cref{fig:bar_plots}).

\subsection{Single-domain alignment}
\begin{figure*}[t]
    \centering
    \includegraphics[trim={0 0 0cm 0},clip,width=1.0\textwidth]{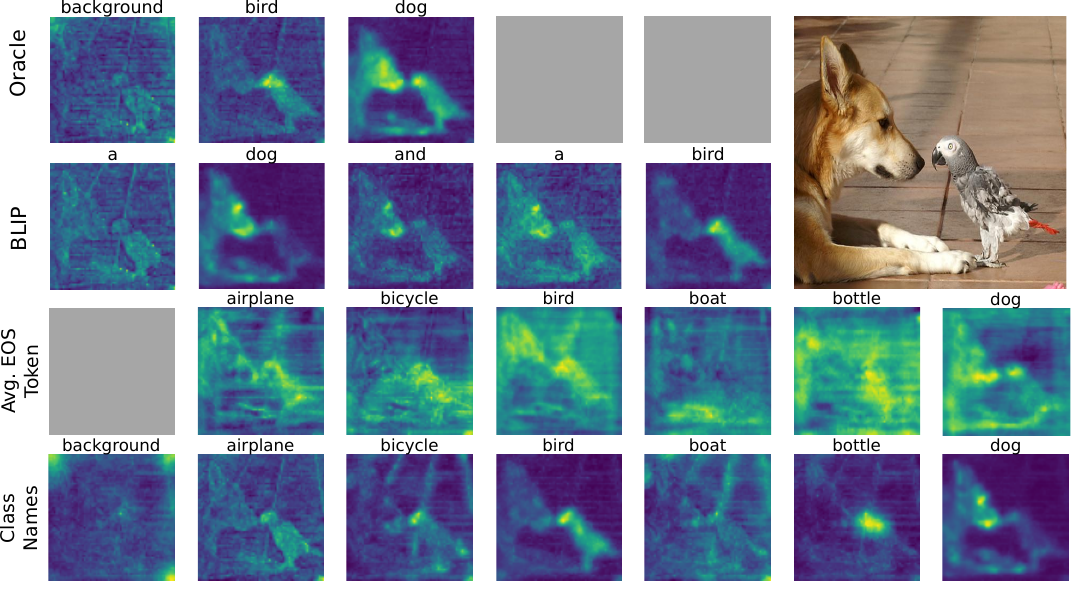}
    \caption{\textbf{Cross-attention maps for different types of prompting (before training).} We compare the cross-attention maps for four types of prompting: oracle, BLIP, Average EOS tokens, and class names as space-separated strings. The cross-attention maps for different heads at all different scales are upsampled to 64x64 and averaged. When comparing Average Template EOS and Class Names, we see (qualitatively) averaging degrades the quality of the cross-attention maps. Furthermore, we find that class names that are not present in the image can have highly localized attention maps (e.g., `bottle'). Further analysis of the cross-attention maps is available in Sec.~\ref{appendix:cross-attention}, where we explore image-to-image generation, copy-paste image modifications, and more. }
    \label{fig:cross_attention_comparison}
\end{figure*}

\textbf{Average EOS Tokens}. We scrutinize the use of average EOS tokens for $\capt$ (see Sec.~\ref{methods_vpd}). While average EOS tokens are sensible when measuring cosine similarities in the CLIP latent space, it is unsuitable in diffusion models, where the text guides the diffusion process through cross-attention. In our qualitative analysis, we find that average EOS tokens degrade the cross-attention maps (\cref{fig:cross_attention_comparison}). 
Instead, we explore using CLIP to embed each class name independently and use the tokens corresponding to the actual word (not the EOS token) and pass this as input to the cross-attention layer: 
\begin{equation}
\captioner_\text{ClassEmbs}(\cnset) = concat(\textit{CLIP}(b) | b \in \cnset) \rightarrow \capt_\text{ClassEmbs} 
\end{equation}
Second, we explore a generic prompt, a string of class names separated by spaces:
\begin{equation}
    \captioner_\text{ClassNames}(\cnset) = \{\text{` '} + b | b \in \cnset\} \rightarrow \capt_\text{ClassNames}
\end{equation}
These prompts are similar to the ones used for averaged EOS tokens $\mathcal{C}_{avg}$ w.r.t. overall text-image alignment but instead use the token corresponding to the word representing the class name. We evaluate these variations on Pascal VOC2012 segmentation. We find that $\capt_\text{ClassNames}$ improves performance by 1.0 mIoU, but $\capt_\text{ClassEmbs}$ reduces performance by 0.3 mIoU (see \cref{tab:pascal}). We perform more in-depth analyses of the effect of text-image alignment on the diffusion model's cross-attention maps and image generation properties in Appendix~\ref{appendix:cross-attention}.

\begin{figure}[t]
    \centering
    \includegraphics[width=\linewidth]{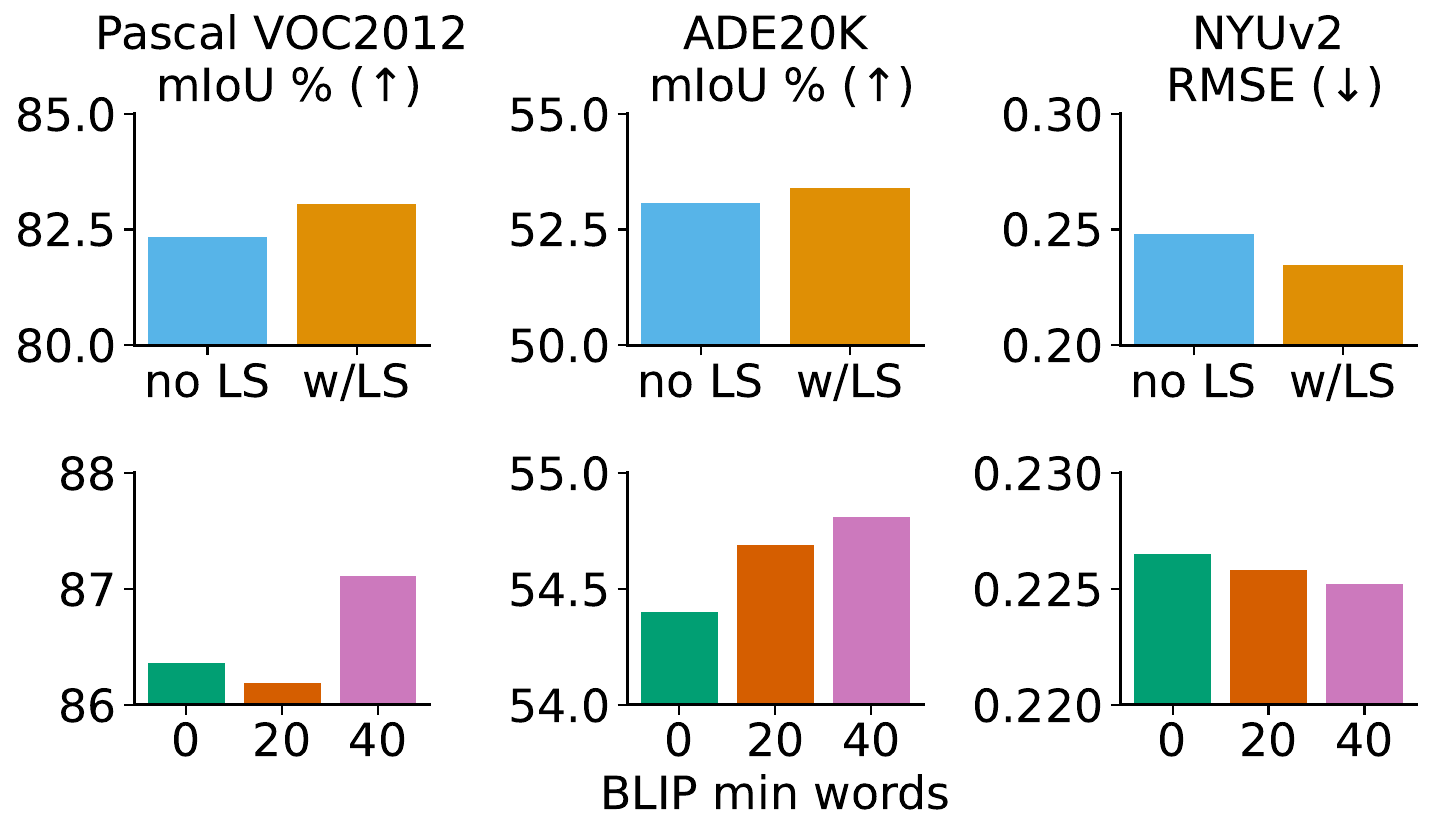}
    \caption{\textbf{Effects of Latent Scaling (LS) and BLIP caption minimum length.} We report mIoU for Pascal, mIoU for ADE20K, and RMSE for NYUv2 depth (right). (Top) Latent scaling improves performance on Pascal ${\sim}0.8$ mIoU (higher is better), ${\sim}0.3$ mIoU, and ${\sim}5.5\%$ relative RMSE (lower is better). (Bottom) We see a similar effect for BLIP minimum token length, with longer captions performing better, improving ${\sim}0.8$ mIoU on Pascal, ${\sim}0.9$ mIoU on ADE20K, and ${\sim}0.6\%$ relative RMSE.}
    \label{fig:bar_plots}
\end{figure}

\textbf{TADP}. To align the diffusion model text input to the image, we use \mbox{BLIP-2} \citep{li_blip-2_2023} to generate captions for every image in our single-domain datasets (Pascal, ADE20K, and NYUv2). 
\begin{equation}    \captioner_{\text{TADP}}(x) = \textit{BLIP-2}(x) \rightarrow \capt_\text{TADP}(x)
\end{equation}
BLIP-2 is trained to produce image-aligned text captions and is designed around the CLIP latent space. However, other vision-language algorithms that produce captions could also be used. We find that these text captions improve performance in all datasets and tasks (Tabs. \ref{tab:pascal}, \ref{tab:ade}, \ref{tab:depth}). Performance improves on Pascal segmentation by $\sim$4\% mIoU, ADE20K by $\sim$1.4\% mIoU, and NYUv2 Depth by a relative RMSE improvement of 4\%. 
We see stronger effects on the fast schedules for ADE20K with an improvement of $\sim$5 mIoU at (4k), $\sim$2.4 mIoU (8K). On NYUv2 Depth, we see a smaller gain on the fast schedule $\sim$2.4\%. 
All numbers are reported relative to VPD with latent scaling.

\input{tables/seg_ade20k_full}
\input{tables/depth_nyu}

We perform some ablations to analyze what aspects of the captions are important. We explore the minimum token number hyperparameter for BLIP-2 to explore if longer captions can produce more useful feature maps for the downstream task. We try a minimum token number of 0, 20, and 40 tokens (denoted as $\capt_\text{TADP-N}$) and find small but consistent gains with longer captions, resulting on average ~0.75\% relative gain for 40 tokens vs. 0 tokens (Fig.~\ref{fig:bar_plots}). Next, we ablate the Pascal $\capt_\text{TADP-20}$ captions to understand what in the caption is necessary for the performance gains we observe. We use NLTK \citep{bird_natural_2009} to filter for the nouns in the captions. In the $\capt_\text{TADP(NO)-20}$ nouns-only caption setting, we achieve 86.4\% mIoU, similar to 86.2\% mIoU with $\capt_\text{TADP-20}$ (\cref{tab:pascal}), suggesting nouns are sufficient.

\textbf{Oracle}. This insight about nouns leads us to ask if an oracle caption, in which all the object class names in an image are provided as a caption, can improve performance further. We define $\cnset(x)$ as the set of class names present in image $x$.
\begin{equation}
    \captioner_{\text{Oracle}}(x) = \{\text{` '} + b | b \in \cnset(x)\} \rightarrow \capt_\text{Oracle}(x)
\end{equation}
While this is not a realistic setting, it serves as an approximate upper bound on performance for our method on the segmentation task. We find a large improvement in performance in segmentation, achieving 89\% mIoU on Pascal and 72.2\% mIoU on ADE20K. For depth estimation, multi-class segmentation masks are only provided for a smaller subset of the images, so we cannot generate a comparable oracle. We perform ablations on the oracle captions to evaluate the model's sensitivity to alignment. For ADE20K, on the 4k iteration schedule, we modify the oracle captions by randomly adding and removing classes such that the recall and precision are at 0.5, 0.75, and 1.0 (independently) (\cref{tab:oracle_ablations}). We find that both precision and recall have an effect, but recall is significantly more important. When recall is lower (0.50), improving precision has minimal impact (\textless1\% mIoU). However, precision has progressively larger impacts as recall increases to 0.75 and 1.00 ($\sim$3\% mIoU and $\sim$7\% mIoU). In contrast, recall has large impacts at every precision level: 0.5 - ($\sim$6\% mIoU), 0.75 - ($\sim$9\% mIoU), and 1.00 - ($\sim$13\% mIoU). BLIP-2 captioning performs similarly to a precision of 1.00 and a recall of 0.5 (\cref{tab:ade}). Additional analyses w.r.t. precision, recall, and object sizes can be found in Appendix~\ref{appendix:additional-results}.

\subsection{Cross-domain alignment}
Next, we ask if text-image alignment can benefit cross-domain tasks. In cross-domain, we train a model on a source domain and test it on a different target domain. There are two aspects of alignment in the cross-domain setting: the first is also present in single-domain, which is image-text alignment; the second is unique to the cross-domain setting, which is text-target domain alignment. The second is challenging because there is a large domain shift between the source and target domain. Our intuition is that while the model has no information on the target domain from the training images, an appropriate text prompt may carry some general information about the target domain. Our cross-domain experiments focus on the text-target domain alignment and use $\captioner_\text{TADP}$ for image-text alignment (following our insights from the single-domain setting).

\textbf{Training.} Our experiments in this setting are designed in the following manner: we train a diffusion model on the source domain captions $\capt_\text{TADP}(x)$. With these source domain captions, we experiment with four different caption modifications (each increasing in alignment to the target domain), a null $\capmod_\textit{null}(\tdinf)$ caption modification where $\capmod_\textit{null}(\tdinf)_s = \varnothing = \capmod_\textit{null}(\tdinf)_{\model} = \varnothing$, a simple $\capmod_\textit{simple}(\tdinf)$ caption modifier where $\capmod_\textit{simple}(\tdinf)_s$ is a hand-crafted string describing the style of the target domain appended to the end and $\capmod_\textit{simple}(\tdinf)_{\model} = \varnothing$, a Textual Inversion \citep{gal_image_2022} $\capmod_\textit{TI}(\tdinf)$ caption modifier where the output $\capmod_\textit{TI}(\tdinf)_s$ is a learned Textual Inversion token $<$*$>$ and $\capmod_\textit{TI}(\tdinf)_{\model} = \varnothing$, and a DreamBooth \citep{ruiz_dreambooth_2022} $\capmod_\textit{DB}(\tdinf)$ caption modifier where $\capmod_\textit{DB}(\tdinf)_s$ is a learned DreamBooth token $<$SKS$>$ and $\capmod_\textit{DB}(\tdinf)_{\model}$ is a DreamBoothed diffusion backbone. We also include two additional control experiments. In the first, $\capmod_\textit{ud}(\tdinf)$ an \textit{unrelated} target domain style is appended to the end of the string. In the second, $\capmod_\textit{nd}(\tdinf)$ a \textit{nearby} but a different target domain style is appended to the caption. $\capmod_\textit{TI}(\tdinf)$ and $\capmod_\textit{DB}(\tdinf)$ require more information than the other methods, such that $\tdinf$ represents a subset of unlabelled images from the target domain.

\input{tables/crossdomain_cityscapes}
\input{tables/crossdomain_pascal}

\textbf{Testing.} When testing the trained models on the target domain images, we want to use the same captioning modification for the test images as in the training setup. However, $\captioner_\text{TADP}$ introduces a confound since it naturally incorporates target domain information. For example, $\captioner_\text{TADP}(x)$ might produce the caption ``a watercolor painting of a dog and a bird" for an image from the Watercolor2K dataset. Using the $\capmod_\textit{simple}(\tdinf)$ captioning modification on this prompt would introduce redundant information and would not match the caption format used during training. In order to remove target domain information and get a plain caption that can be modified in the same manner as in the training data, we use GPT-3.5 \citep{brown2020language} to remove all mentions of the target domain shift. For example, after using GPT-3.5 to remove mentions of the watercolor style in the above sentence, we are left with ``an image of a bird and a dog''. With these \textit{\mbox{GPT-3.5} cleaned captions}, we can match the caption modifications used during training when evaluating test images. This caption-cleaning strategy lets us control how target domain information is included in the test image captions, ensuring that test captions are in the same domain as train captions.

\textbf{Evaluation.}
We evaluate cross-domain transfer on several datasets. We train our model on Pascal VOC \citep{pascal-voc-2007, everingham_pascal_2012} object detection and evaluate on Watercolor2K (W2K) \citep{inoue_cross-domain_2018} and Comic2K (C2K) \citep{inoue_cross-domain_2018}. We also train our model on the Cityscapes \citep{cordts_cityscapes_2016} dataset and evaluate on the Nighttime Driving (ND) \citep{dai_dark_2018} and Dark Zurich-val (DZ-val) \citep{sakaridis_guided_2019} datasets. We show results in Tabs.~\ref{tab:cross-domain-cityscapes},~\ref{tab:cross-domain-pascal}. In the following sections, we also report the average performance of each method on the cross-domain segmentation datasets (average mIoU) and the cross-domain object detection datasets (average AP).

\textbf{Null caption modifier.}
The null captions have no target domain information. In this setting, the model is trained with captions with no target domain information and tested with GPT-3.5 cleaned target domain captions. We find diffusion pre-training to be extraordinarily powerful on its own, with just plain captions (no target domain information); the model already achieves SOTA on VOC$\to$W2K with 72.1 $AP_{50}$, SOTA on CD$\to$DZ-val with 42.8 mIoU and SOTA on CD$\to$ND with 60.8 mIoU. Our model performs better than the current SOTA \citep{topal2022domain} on VOC$\to$W2K and worse on VOC$\to$C2K (highlighted in yellow in \cref{tab:cross-domain-pascal}). However, \cite{topal2022domain} uses a large extra training dataset from the target (comic) domain, so we highlight in bold our results in \cref{tab:cross-domain-pascal} to show they outperform all other methods that use only images in C2K as examples from the target domain. Furthermore, these results are with a lightweight FPN \citep{kirillov_panoptic_2019} head, in contrast to other competitive methods like Refign \citep{bruggemann_refign_2022}, which uses a heavier decoder head.
These captions achieve 50.5 average mIoU and 36.6 average AP.

\textbf{Simple caption modifier.}
We then add target domain information to our captions by prepending the target domain's semantic shift to the generic captions. These caption modifiers are hand-crafted. For example, ``a dog and a bird'' becomes ``a X style painting of a dog and a bird'' (where X is watercolor for W2K and comic for C2K) and ``a dark night photo of a dog and a bird'' for DZ. These captions achieve 48.0 average mIoU and 37.7 average AP. 

\textbf{Textual Inversion caption modifier.}
Textual inversion \citep{gal_image_2022} is a method that learns a target concept (an object or style) from a set of images and encodes it into a new token. We learn a novel token from target domain image samples to further increase image-text alignment (for details, see \cref{appendix:model_personalization}). In this setting, the sentence template becomes ``a $<$token$>$ style painting of a dog and a bird''. We find that, on average, Textual Inversion captions perform the best, achieving 51.1 average mIoU and 38.2 average AP. 

\textbf{DreamBooth caption modifier.}
DreamBooth-ing \citep{ruiz_dreambooth_2022} aims to achieve the same goal as textual inversion. Along with learning a new token, the stable-diffusion backbone itself is fine-tuned with a set of target domain images (for details, see \cref{appendix:model_personalization}). We swap the stable diffusion backbone with the DreamBooth-ed backbone before training. We use the same template as in textual inversion. These captions achieve 49.7 average mIoU and 38.1 average AP. 

\textbf{Ablations.}
We ablate our target domain alignment strategy by introducing unrelated and nearby target-domain style modifications. For example, this would be ``a \textbf{dashcam} photo of a dog and a bird" (unrelated) and ``a \textbf{constructivism} painting of a dog and a bird" (nearby) for the W2K  and C2K datasets. ``A \textbf{watercolor} painting of a car on the street" (unrelated) and  ``a \textbf{foggy} photo of a car on the street" for the ND and DZ-val datasets. We find these off-target domains reduce performance on all datasets. 

\section{Discussion}

We present a method for image-text alignment that is general, fully automated, and can be applied to any diffusion-based perception model. To achieve this, we systematically explore the impact of text-image alignment on semantic segmentation, depth estimation, and object detection. We investigate whether similar principles apply in the cross-domain setting and find that alignment towards the target domain during training improves downstream cross-domain performance. 

We find that EOS token averaging for prompting does not work as effectively as strings for the objects in the image. Our oracle ablation experiments show that our diffusion pre-trained segmentation model is particularly sensitive to missing classes (reduced recall) and less sensitive to off-target classes (reduced precision), and both have a negative impact. Our results show that aligning text prompts to the image is important in identifying/generating good multi-scale feature maps for the downstream segmentation head. This implies that the multi-scale features and latent representations do not naturally identify semantic concepts without the guidance of the text in diffusion models. Moreover, proper latent scaling is crucial for downstream vision tasks. Lastly, we show how using a captioner, which has the benefit of being open vocabulary, high precision, and downstream task agnostic, to prompt the diffusion pre-trained segmentation model automatically improves performance significantly over providing all possible class names.

We also find that diffusion models can be used effectively for cross-domain tasks. Our model, without any captions, already surpasses several SOTA results in cross-domain tasks due to the diffusion backbone's generalizability. We find that good target domain alignment can help with cross-domain performance for some domains, and misalignment leads to worse performance. Capturing information about target domain styles in words alone can be difficult. For these cases, we show that model personalization through Textual Inversion or Dreambooth can bridge the gap without requiring labeled data. Future work could explore how to expand our framework to generalize to multiple unseen domains.
%
Future work may also explore closed vocabulary captioners that are more task-specific to get closer to oracle-level performance. 

{
    \small
\paragraph{Acknowledgements.}
Pietro Perona and Markus Marks were supported by the National Institutes of Health (NIH R01 MH123612A) and the Caltech Chen Institute (Neuroscience Research Grant Award). Pietro Perona, Neehar Kondapaneni, Rogerio Guimaraes, and Markus Marks were supported by the Simons Foundation (NC-GB-CULM-00002953-02).
Manuel Knott was supported by an ETH Zurich Doc.Mobility Fellowship. 
We thank Oisin Mac Aodha, Yisong Yue, and Mathieu Salzmann for their valuable inputs that helped improve this work. 
}

{
    \small
    \bibliographystyle{ieeenat_fullname}
    \bibliography{references,reference_manual}
}
\newpage
\clearpage
\setcounter{page}{1}
\beginsupplement
\onecolumn

\begin{center}
    \begin{minipage}{\textwidth}
        \centering
        \fontsize{16}{20}\selectfont
        \textbf{\thetitle\\Supplementary Materials}
    \end{minipage}
\end{center}
\vspace{12pt}
\section{Cross-attention analysis}
\label{appendix:cross-attention}
\begin{figure*}[ht!]
    \centering
    \includegraphics[trim={0 20cm 0 0cm}, clip, width=1\textwidth]{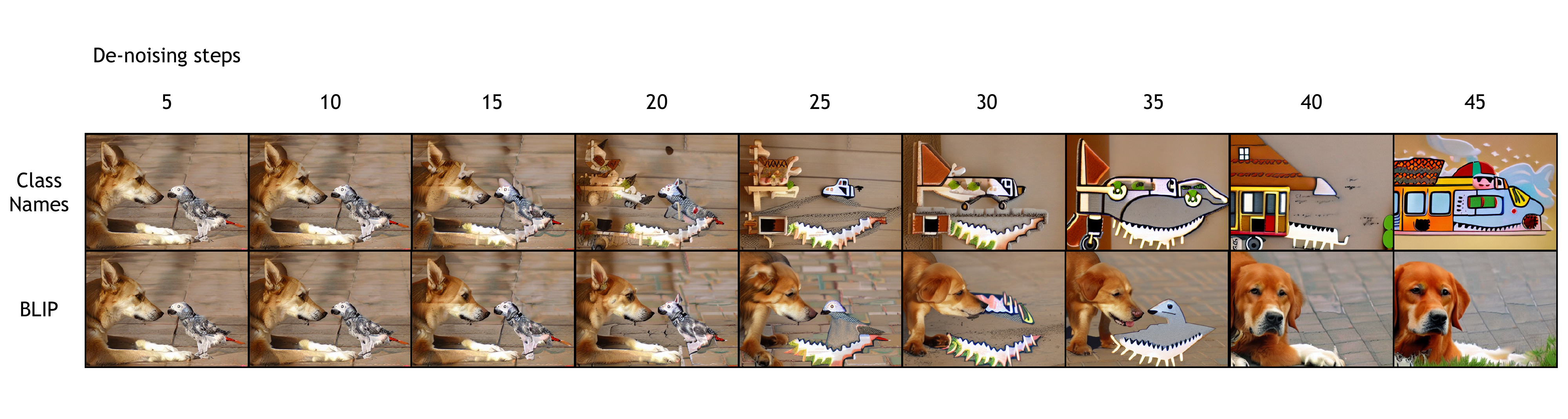}
    \caption{\textbf{Qualitative image-to-image variation.} An untrained stable diffusion model is passed an image to perform image-to-image variation. The number of denoising steps conducted increases from left to right (5 to 45 out of a total of 50). On the top row, we pass all the class names in Pascal VOC 2012: \textit{``background airplane bicycle bird boat bottle bus car cat chair cow dining table dog horse motorcycle person potted plant sheep sofa train television''}. In the bottom row we pass the BLIP caption \textit{``a bird and a dog''}.}
    \label{fig:img2img}
\end{figure*}

\begin{figure*}[ht!]
    \centering
    \includegraphics[width=1\linewidth]{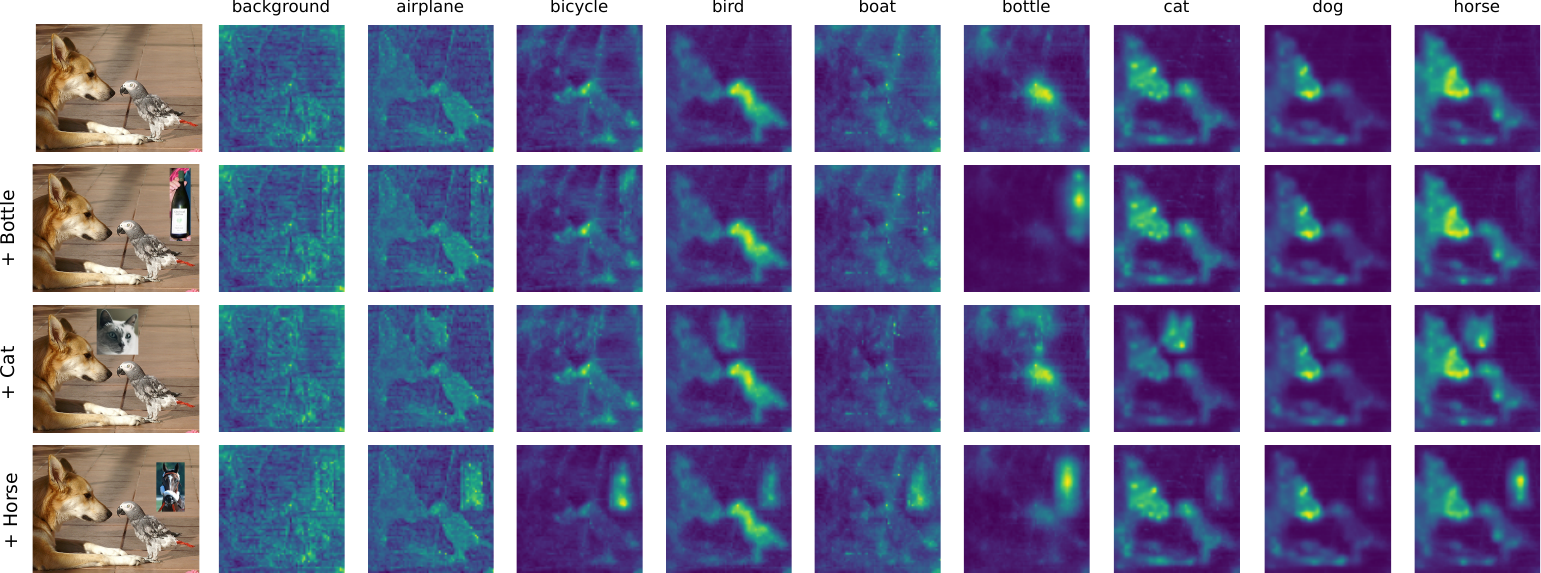}
    \caption{\textbf{Copy-Paste Experiment.} A bottle, a cat, and a horse from different images are copied and pasted into our base image to see how the cross-attention maps change. The label on the left describes the category of the item that has been pasted into the image. The labels above each map describe the cross-attention map corresponding to the token for that label.} 
    \label{fig:copy_paste}
\end{figure*}

\begin{figure*}[p]
    \centering
    \begin{subfigure}[b]{0.48\textwidth}
    \centering
    \includegraphics[width=1\textwidth]{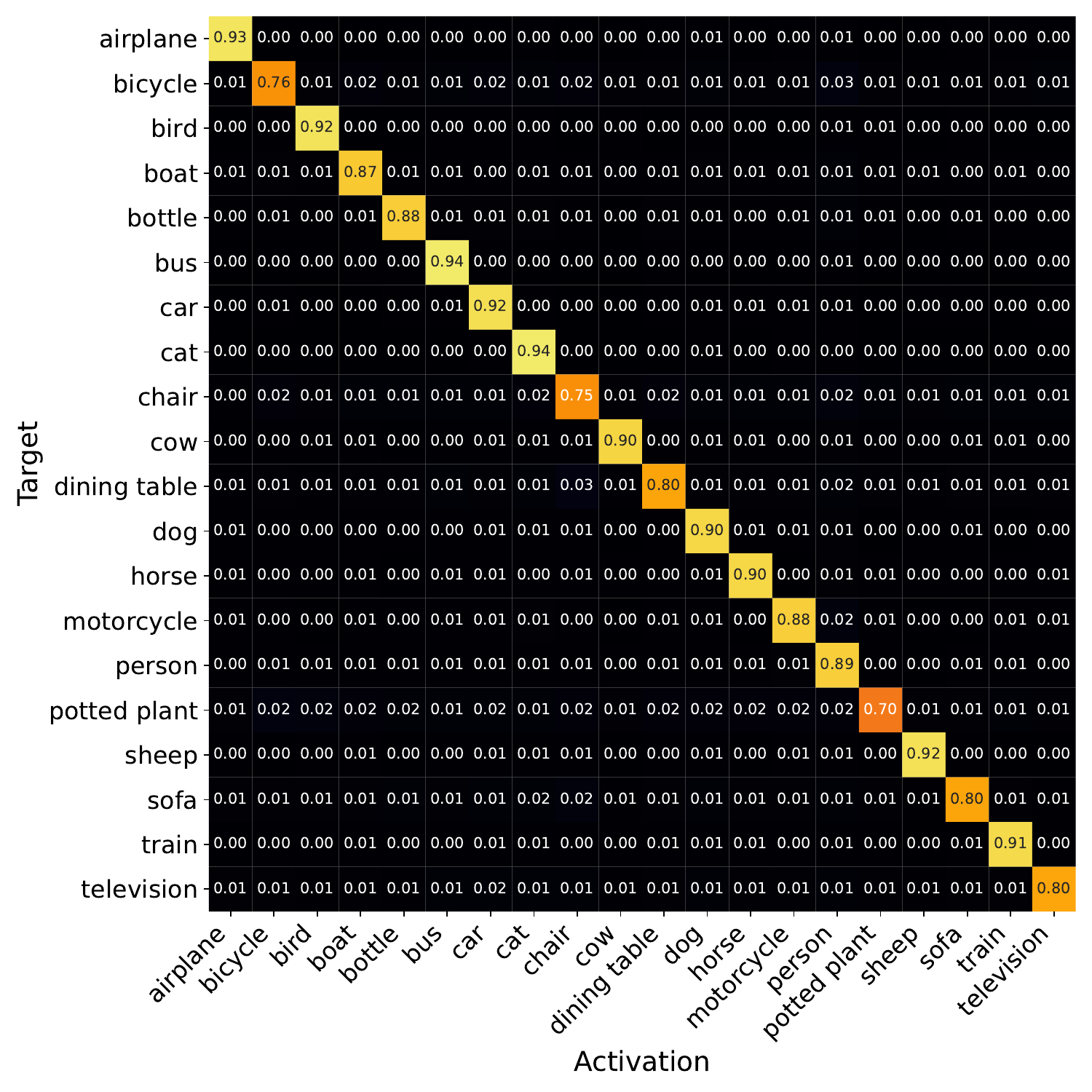}
    \end{subfigure}
    \begin{subfigure}[b]{0.48\textwidth}
    \centering
    \includegraphics[width=\textwidth]{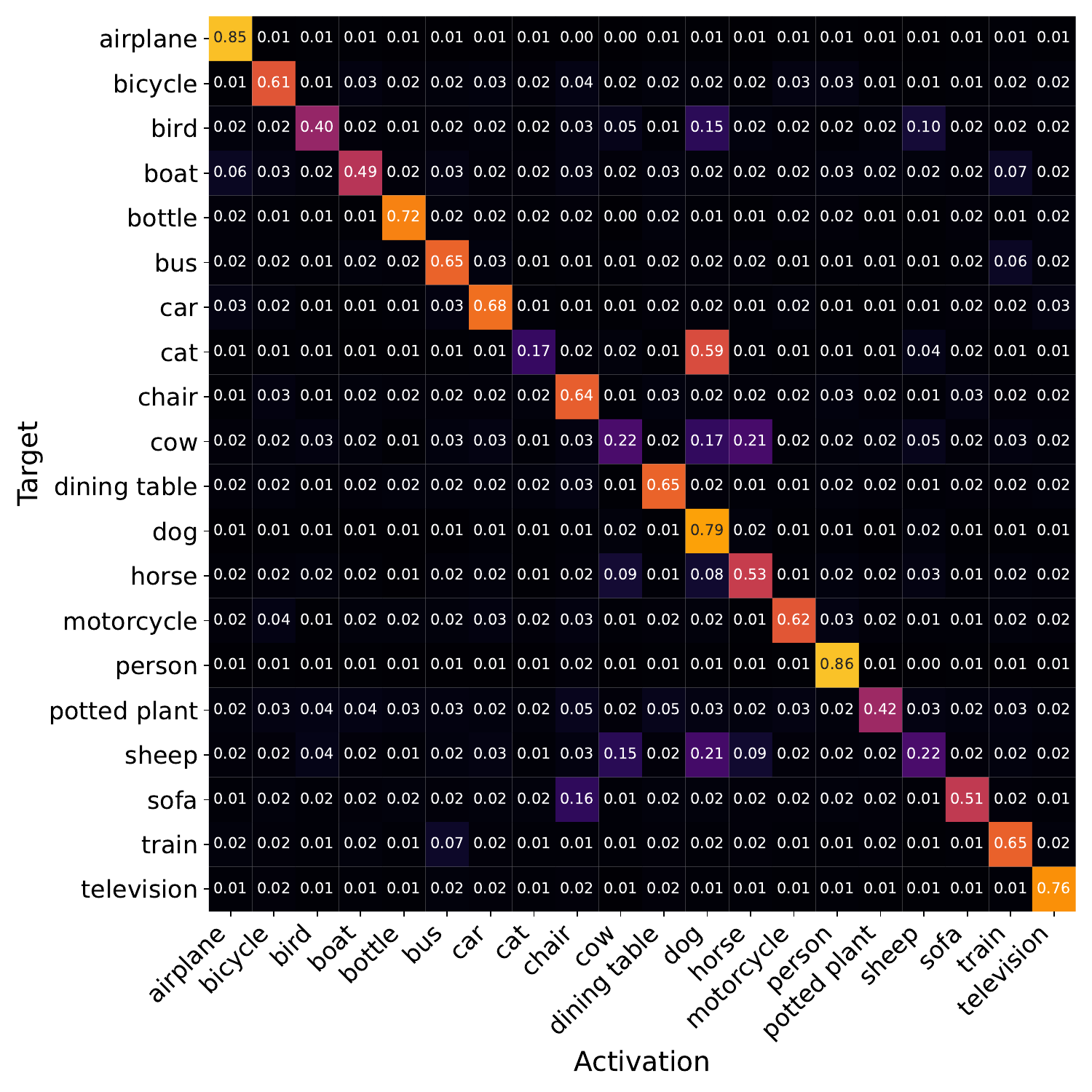}
    \end{subfigure}
    \caption{\textbf{Impact of off-target classes on semantic segmentation performance.} The matrices show normalized scores averaged over pixels on Pascal VOC 2012 for an oracle-trained model when receiving either present class names (left) or all class names (right).}
    \label{fig:px-confusion}
\end{figure*}

\textbf{Qualitative image-to-image variation analysis.}
We present a qualitative and quantitative analysis of the effect of off-target class names added to the prompt. In \cref{fig:img2img}, we use the stable diffusion image to image (img2img) variation pipeline (with the original Stable Diffusion 1.5 weights) to qualitatively analyze the effects of prompts with off-target classes. The img2img variation pipeline encodes a real image into a latent representation, adds a user-specified amount of noise to the latent representation, and de-noises it (according to a user-specified prompt) to generate a variation on the original image. The amount of noise added is dictated by a strength ratio indicating how much variation should occur. A higher ratio results in more added noise and more denoising steps, allowing a relatively higher impact of the new text prompt on the image. We find that $\capt_{\text{ClassNames}}$ (see caption for details) results in variations that incorporate the off-target classes. This effect is most clear looking across the panels left to right in which objects belonging to off-target classes (an airplane and a train) become more prominent. These qualitative results imply that this prompt modifies the latent representation to incorporate information about off-target classes, potentially making the downstream task more difficult. In contrast, using the BLIP prompt changes the image, but the semantics (position of objects, classes present) of the image variation are significantly closer to the original. These results suggest a mechanism for how off-target classes may impact our vision models. We quantitatively measure this effect using a fully trained Oracle model in the following section.
\\
\\\textbf{Copy-Paste Experiment.} An interesting property in Fig.~\ref{fig:cross_attention_comparison} is that the word bottle has strong cross-attention over the neck of the bird. We hypothesize that diffusion models seek to find the nearest match for each token since they are trained to generate images that correspond to the prompt. We test this hypothesis on a base image of a dog and a bird. We first visualize the cross-attention maps for a set of object labels. We find that the words bottle, cat, and horse have a strong cross-attention to the bird, dog, and dog, respectively. We paste a bottle, cat, and horse into the base image to see if the diffusion model will localize the ``correct'' objects if they are present. In Fig.~\ref{fig:copy_paste}, we show that the cross-attention maps prefer to localize the ``correct'' object, suggesting our hypothesis is correct.
\\
\\\textbf{Averaged EOS Tokens: Averaging vs. EOS?}
Averaged EOS Tokens create diffuse attention maps that empirically harm performance. What is the actual cause of the decrease in performance? Is it averaging, or is it the usage of many EOS tokens? We replace the averaged EOS tokens with single prompt EOS tokens and find that the attention maps are still diffuse. This indicates that the usage of EOS tokens is the primary cause of the diffuse attention maps and not the averaging. 
\\\textbf{Quantitative effect of $\capt_{\text{ClassNames}}$ on Oracle model.} To quantify the impact of the off-target classes on the downstream vision task, we measure the averaged pixel-wise scores (normalized via Softmax) per class when passing the $\capt_{\text{ClassNames}}$ to the Oracle segmentation model for Pascal VOC 2012 (\cref{fig:px-confusion}). We compare this to the original oracle prompt. We find that including the off-target prompts significantly increases the probability of a pixel being misclassified as one of the semantically nearby off-target classes. For example, if the original image contains a cow, including the words dog and sheep, it significantly raises the probability of misclassifying the pixels belonging to the cow as pixels belonging to a dog or a sheep. These results indicate that the task-specific head picks up the effect of off-target classes and is incorporated into the output. 

\begin{figure*}[ht!]
    \centering
    \includegraphics[width=1\linewidth]{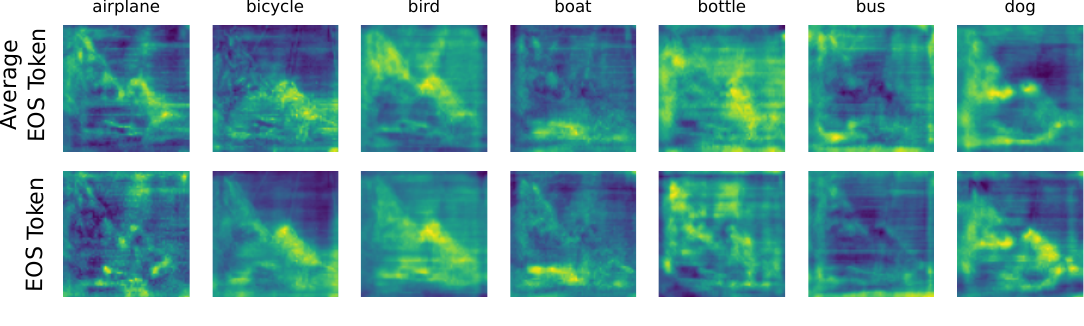}
    \caption{\textbf{Averaging vs. EOS.} In ~\cite{zhao_unleashing_2023}, for each class name, the EOS token from 80 prompts (containing the class name) was averaged together. The averaged EOS tokens for each class were concatenated together and passed to the diffusion model as text input. We explore if averaging drives the diffuse nature of the cross-attention maps. We replace the 80 prompt templates with a single prompt template: ``a photo of a \{class name\}'' and visualize the cross-attention maps. In the top row, we show the averaged template EOS tokens. In the bottom row, we show the single template EOS tokens. } 
    \label{fig:average_vs_single}
\end{figure*}

\clearpage

\section{Additional ADE20K Results}
\label{appendix:additional-results}
\vspace{20pt}
\begin{table}[ht!]
  \centering

  \adjustbox{width=0.5\linewidth}{
    \begin{tabular}{lcccc}
    \toprule
    \multirow{2}{*}{Method} & \multicolumn{2}{c}{4K Iters} & \multicolumn{2}{c}{8K Iters} \\ \cmidrule(lr){2-3} \cmidrule(lr){4-5}
          & mIoU$^{\rm ss}$ & mIoU$^{\rm ms}$ & mIoU$^{\rm ss}$ & mIoU$^{\rm ms}$ \\ \midrule
    VPD (null text) & 41.5 & - & 46.9 & - \\
    VPD$_{\rm A32}$ \citep{zhao_unleashing_2023} & 43.1 & 44.2 & 48.7 & 49.5  \\
    VPD(R)          & 42.6 & 43.6 & 49.2 & 50.4 \\
    VPD(LS)        & 45.0 & 45.8 & 50.5 & 51.1 \\
    \rowcolor{CBLightGreen} \ours-20 (Ours)         & \textbf{50.2} & \textbf{50.9} & \textbf{52.8} & \textbf{54.1} \\
    \rowcolor{CBLightGreen}\ours(TA)-20 (Ours)    & 49.9 & 50.7 & 52.7 & 53.4 \\
    \bottomrule
    \end{tabular}
  }
  \caption{\textbf{Semantic segmentation fast schedule on ADE20K.} Our method has a large advantage over prior work on the fast schedule with significantly better performance in both the single-scale and multi-scale evaluations for 4k and 8k iterations.}
  \label{tab:seg_ade_fast}
\end{table}
\vspace{20pt}
\input{tables/oracle_ablations}

\begin{figure}[ht]
  \centering
  \begin{minipage}[t]{0.45\linewidth}
\includegraphics[width=\textwidth]{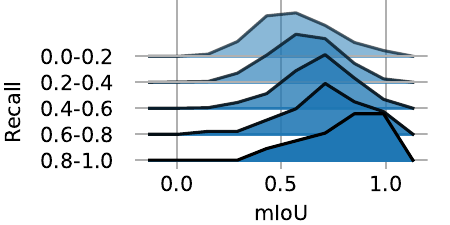}
    \caption{\textbf{Recall analysis.} ADE20k mIOU per image with respect to the recall of classes present in the caption. We embedded each word in our caption with CLIP's text encoder. We considered a cosine similarity of $\geq0.9$ with the embedded class name as a match. Linear regression analysis shows positive correlations between recall and mIoU ($r=0.28$).}
    \label{fig:recall}
  \end{minipage}
  \hfill
  \begin{minipage}[t]{0.45\linewidth}
\includegraphics[width=\textwidth]{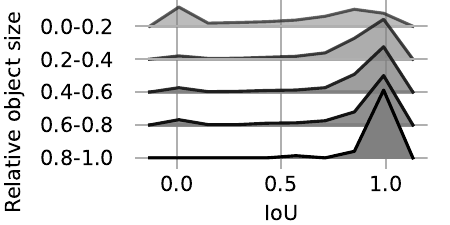}
\caption{\textbf{Object size analysis.} ADE20k IOU per object image with respect to the relative object size (pixels divided by total pixels). Linear regression analysis shows positive correlations between relative object size and the IoU-score of a class ($r=0.40$).}
\label{fig:object-size}
  \end{minipage}
\end{figure}

\newpage
\clearpage

\section{Qualitative Examples}
\vspace{-10pt}
\begin{figure}[h]
    \centering
    \includegraphics[width=1.0\textwidth]{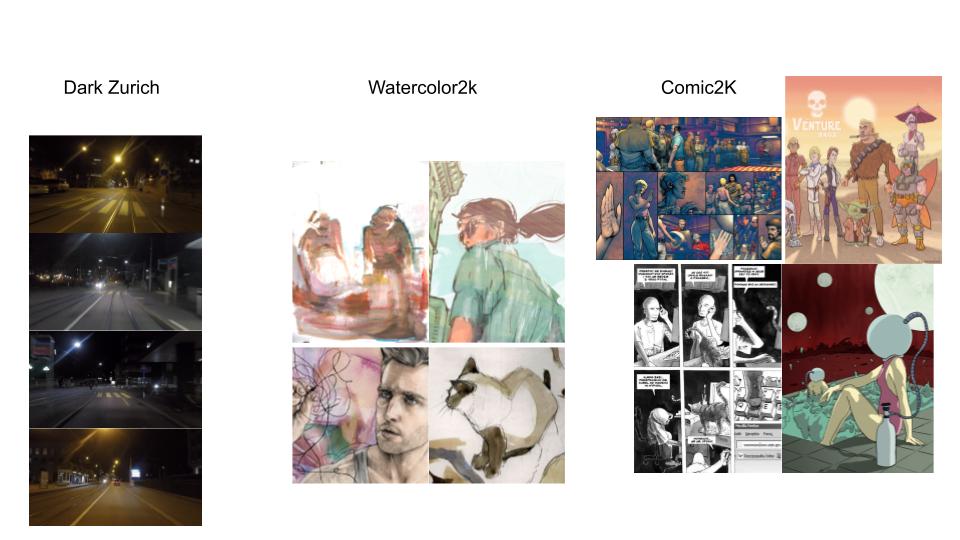}
    \caption{\textbf{Ground truth examples of the tokenized datasets.}}
\end{figure}
\begin{figure}[h]
    \centering
    \includegraphics[width=1.0\textwidth]{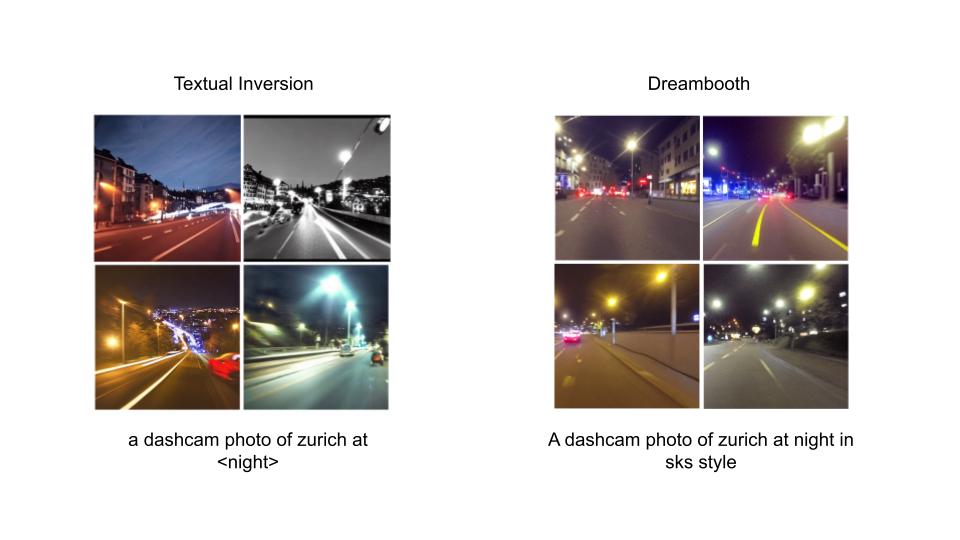}
    \vspace{-1cm}
    \caption{\textbf{Textual inversion and Dreambooth tokens of Cityscapes to Dark Zurich.}}
\end{figure}
\vspace{25pt}
\begin{figure}[h]
    \centering
    \includegraphics[width=1.0\textwidth]{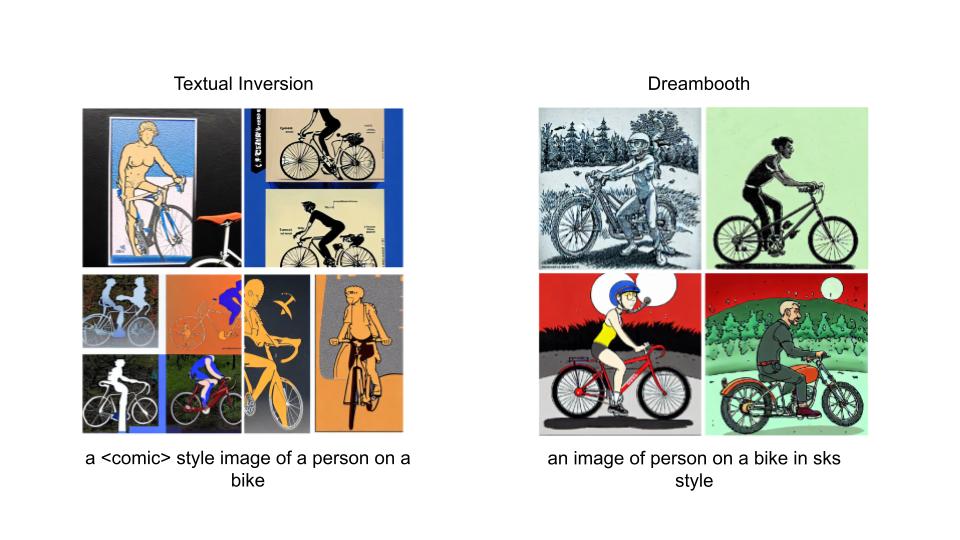}
    \vspace{-1cm}
    \caption{\textbf{Textual inversion and Dreambooth tokens of VOC to Comic.}}
\end{figure}
\begin{figure}
    \centering
    \includegraphics[width=1.0\textwidth]{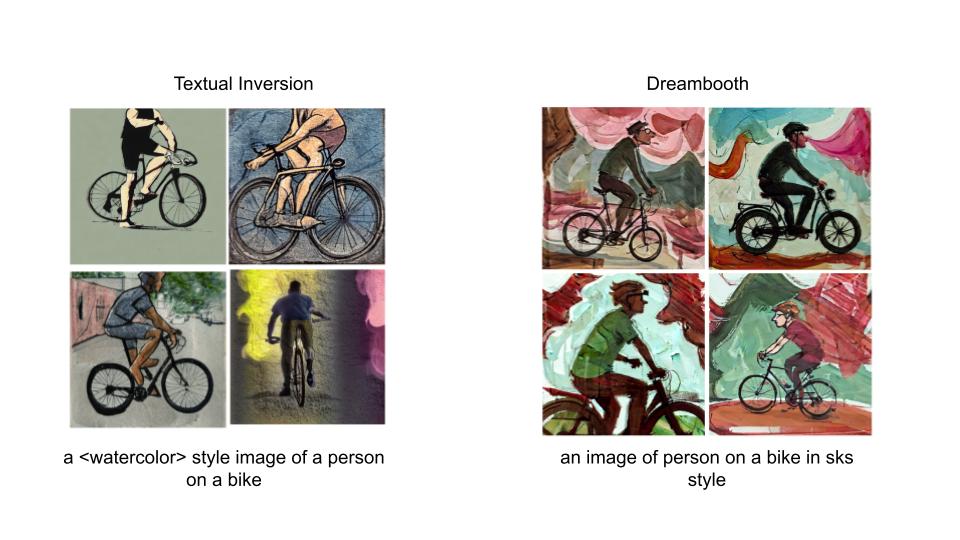}
    \vspace{-1cm}
    \caption{\textbf{Textual inversion and Dreambooth tokens of VOC to Watercolor.}}
\end{figure}
\newpage
\clearpage

\begin{figure*}[h]
    \centering
    \includegraphics[width=\textwidth]{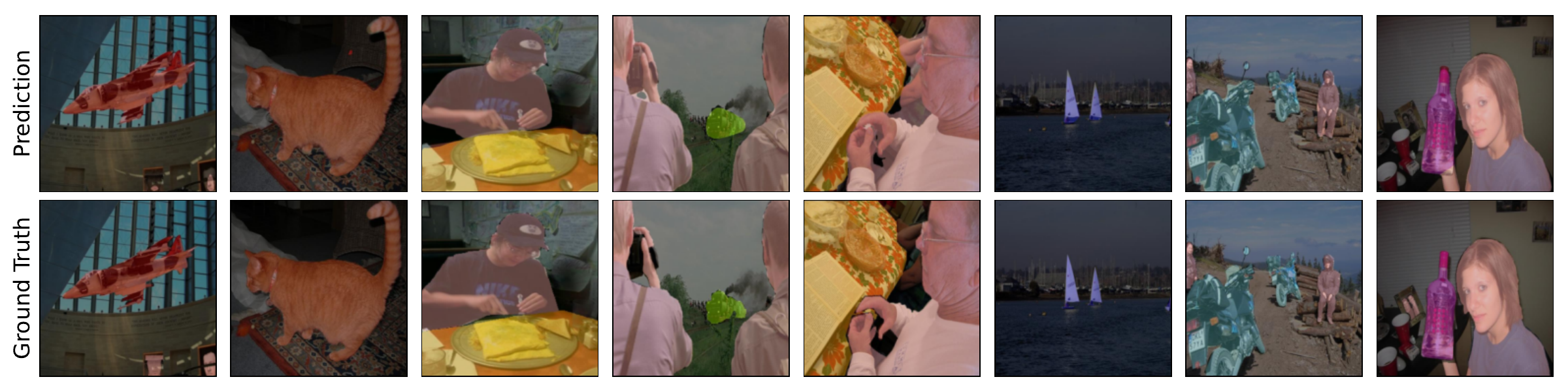}
    \caption{\textbf{Predictions (top) and Ground Truth (bottom) visualizations for Pascal VOC2012.}}
\end{figure*}

\vspace{75pt}
\begin{figure*}[h]
    \centering
    \includegraphics[width=\textwidth]{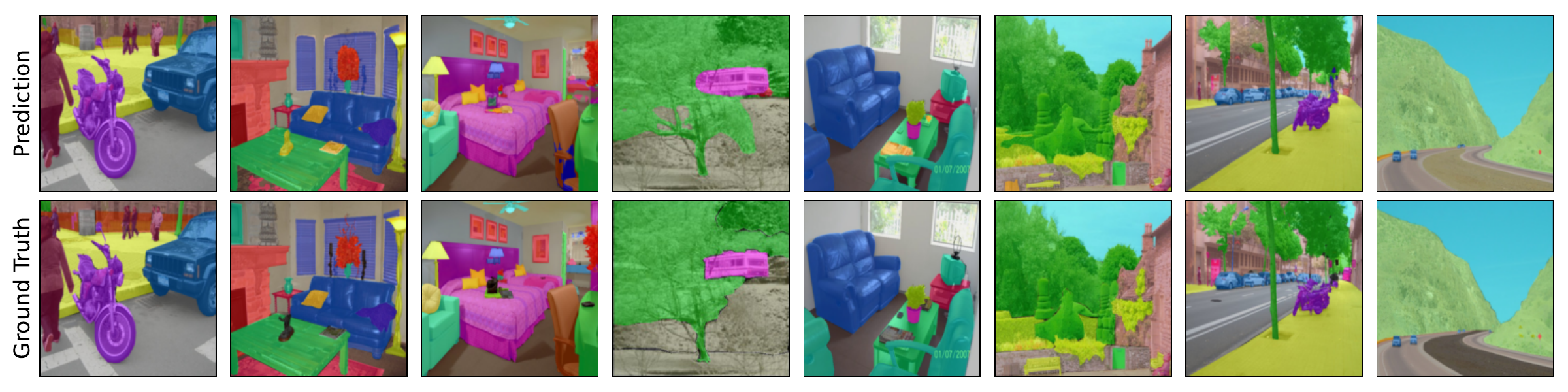}
    \caption{\textbf{Predictions (top) and Ground Truth (bottom) visualizations for ADE20K.}}
\end{figure*}

\vspace{75pt}
\begin{figure*}[h]
    \centering 
    \includegraphics[width=\textwidth]{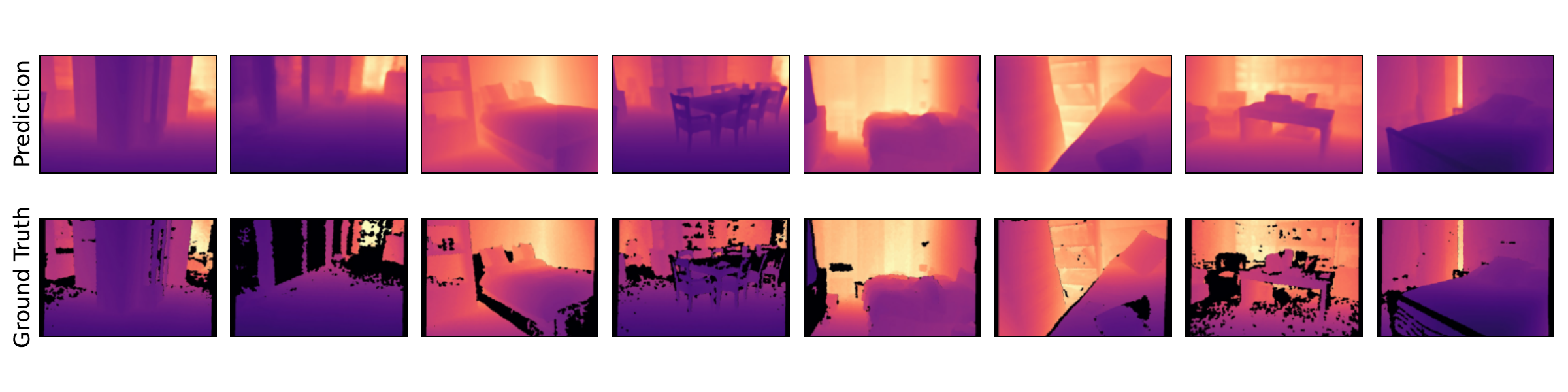}
    \caption{\textbf{{Predictions (top) and Ground Truth (bottom) visualizations for NYUv2 Depth.}}}
\end{figure*}

\newpage
\clearpage

\vspace{75pt}
\begin{figure*}[h]
    \centering 
    \includegraphics[width=\textwidth]{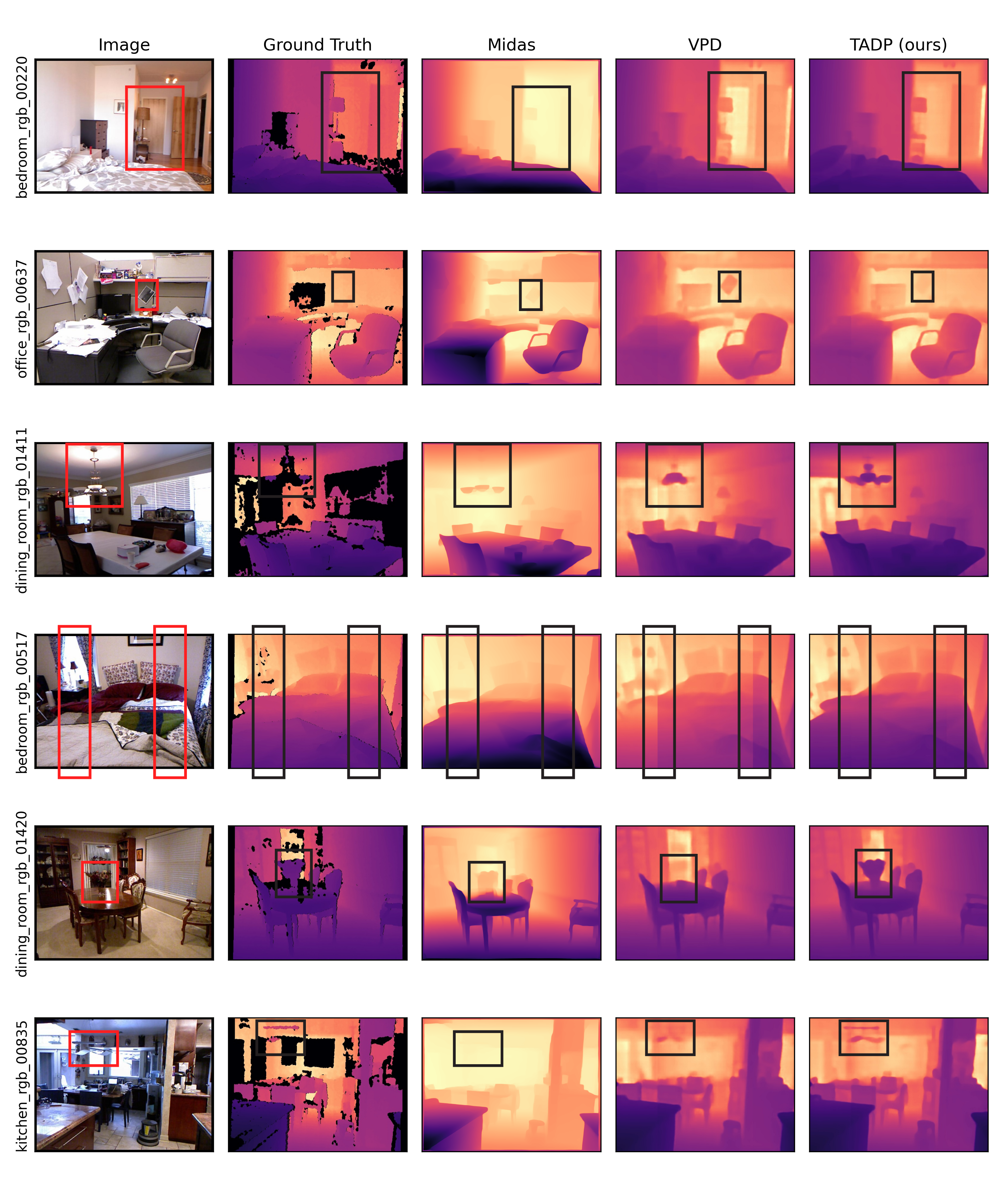}
    \caption{\textbf{Depth Estimation Comparison: Image, Ground Truth, and Prediction visualizations for Midas, VPD, and TADP (ours) in NYUv2 Depth.}
    Black boxes (red on original image) show where TADP is better than Midas and/or VPD.}
\end{figure*}

\vspace{75pt}
\begin{figure*}[h]
    \centering 
    \includegraphics[width=\textwidth]{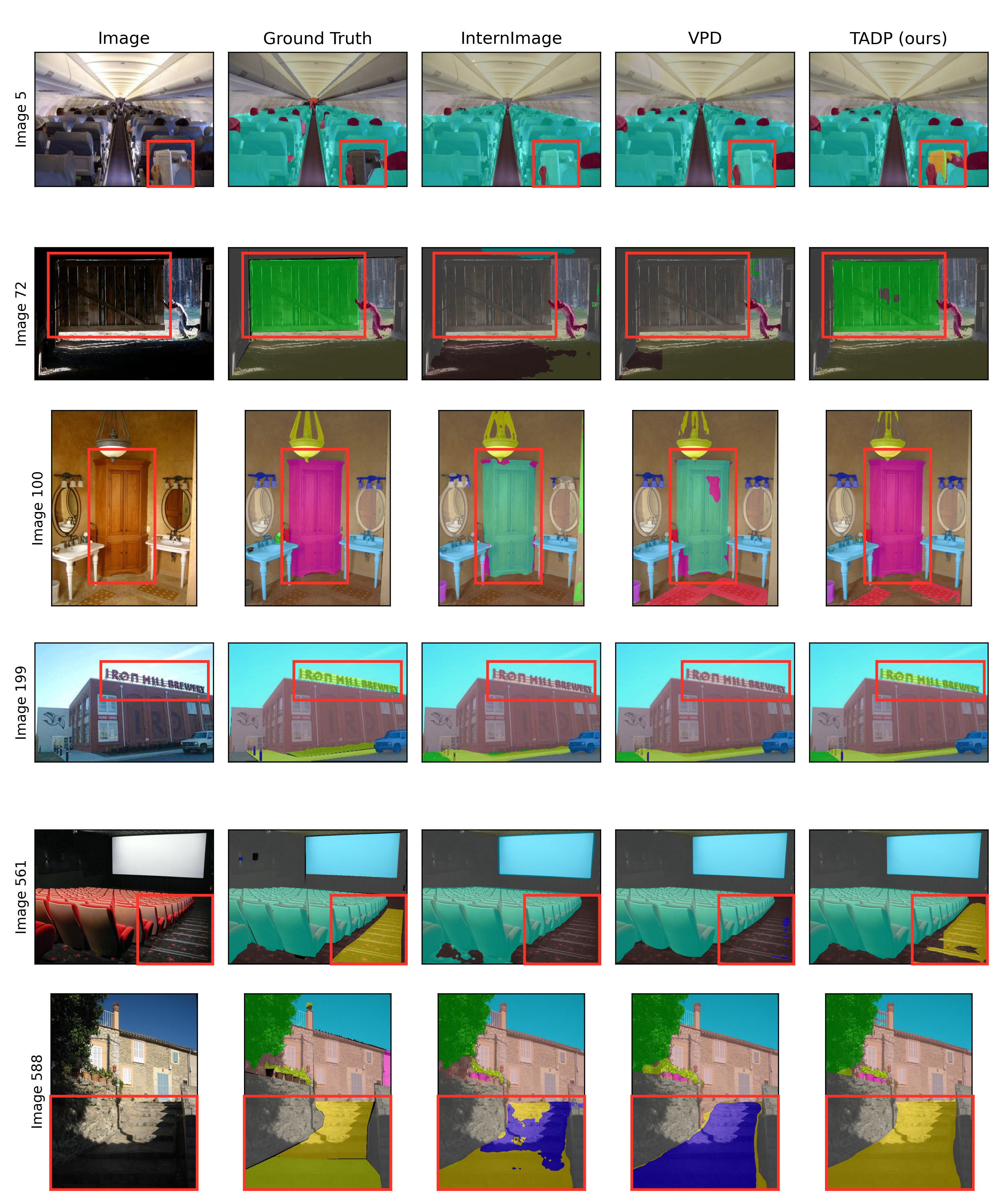}
    \caption{\textbf{Image Segmentation Comparison: Image, Ground Truth, and Prediction visualizations for InternImage, VPD, and TADP (ours) in ADE20K.}
    Red boxes show where TADP is better than InternImage and/or VPD.}
\end{figure*}

\vspace{75pt}
\begin{figure*}[h]
    \centering 
    \includegraphics[width=\textwidth]{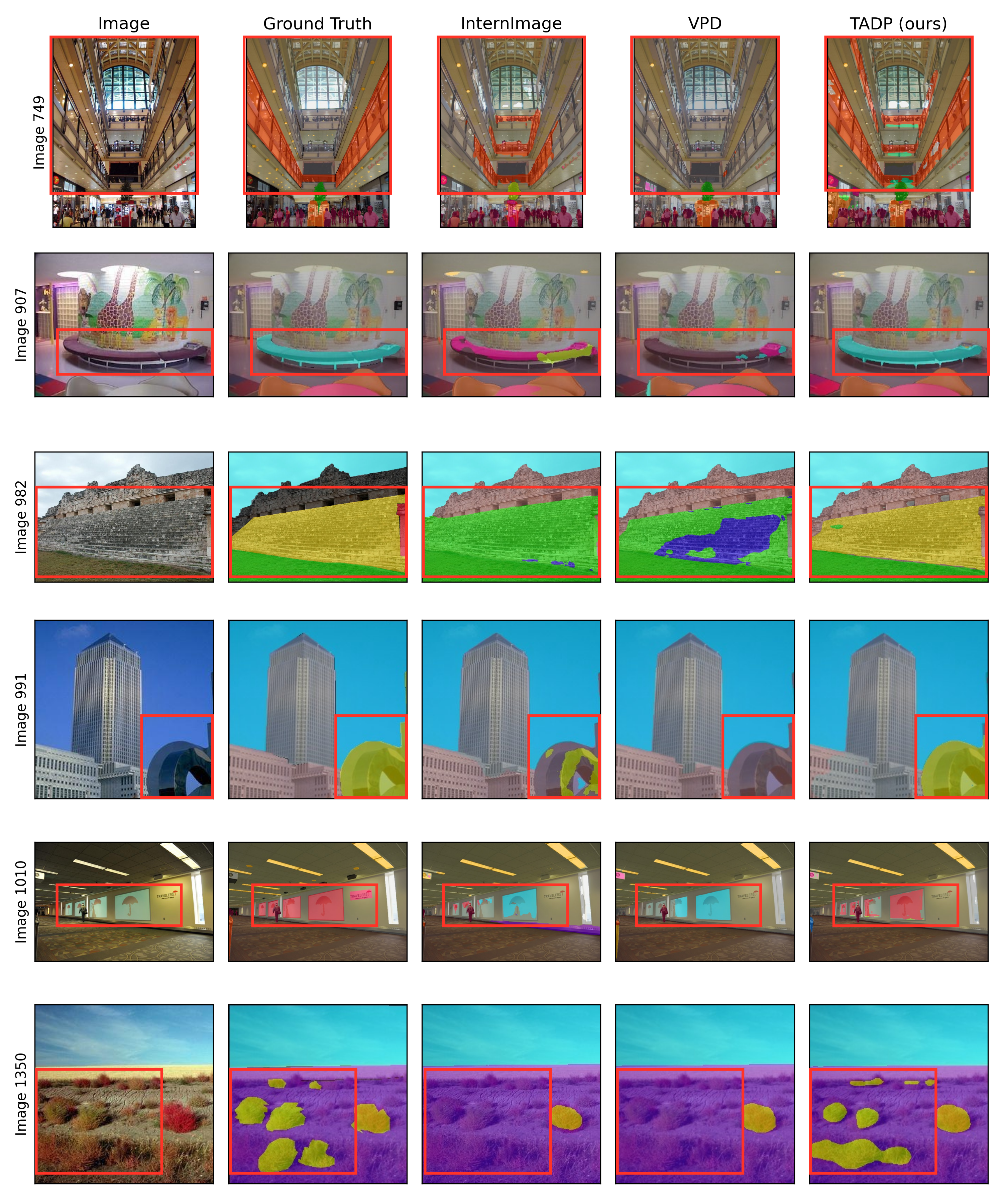}
    \caption{\textbf{Image Segmentation Comparison: Image, Ground Truth, and Prediction visualizations for InternImage, VPD, and TADP (ours) in ADE20K.}
    Red boxes show where TADP is better than InternImage and/or VPD.}
\end{figure*}

\newpage
\clearpage

\vspace{75pt}
\begin{figure*}[h]
    \centering 
    \includegraphics[width=0.85\textwidth]{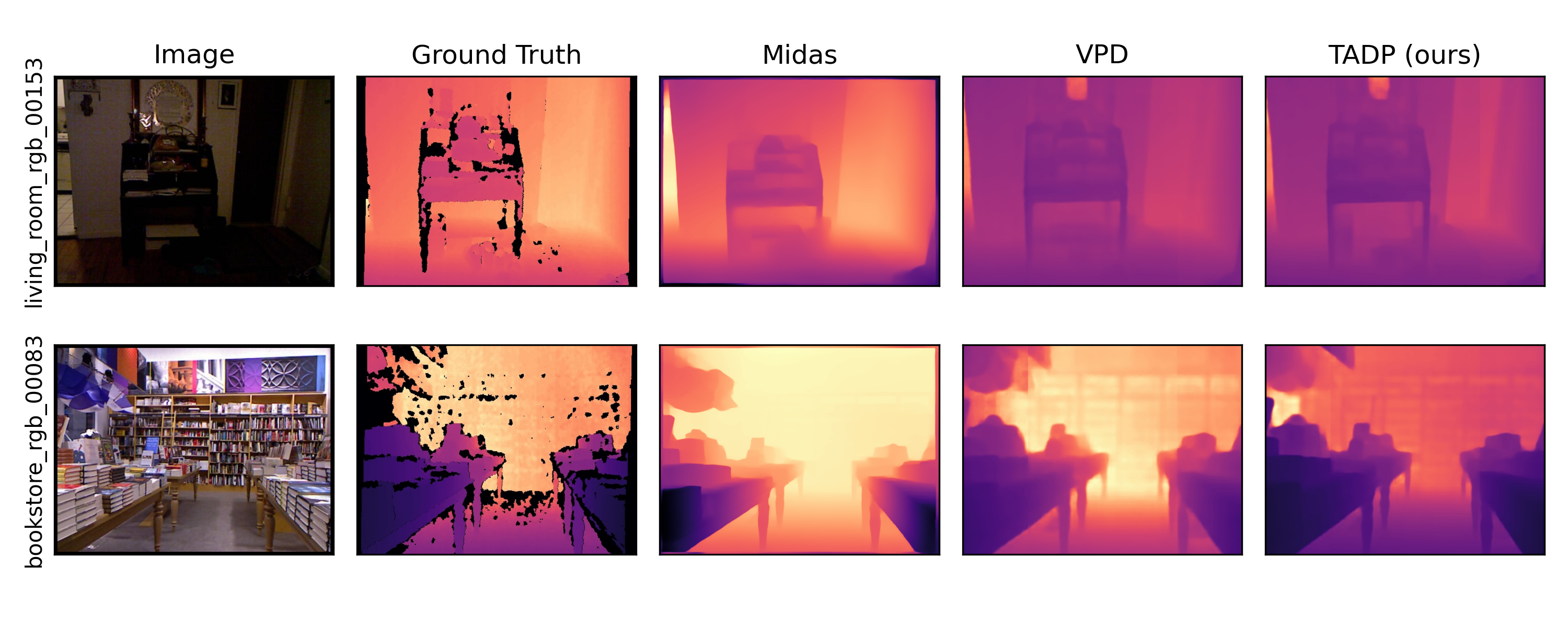}
    \caption{\textbf{Depth Estimation Comparison: Image, Ground Truth, and Prediction visualizations for Midas, VPD, and TADP (ours) in NYUv2 Depth.}
    TADP is worse than Midas and/or VPD in these images in terms of the general scale}
\end{figure*}

\begin{figure*}[h]
    \centering 
    \includegraphics[width=0.85\textwidth]{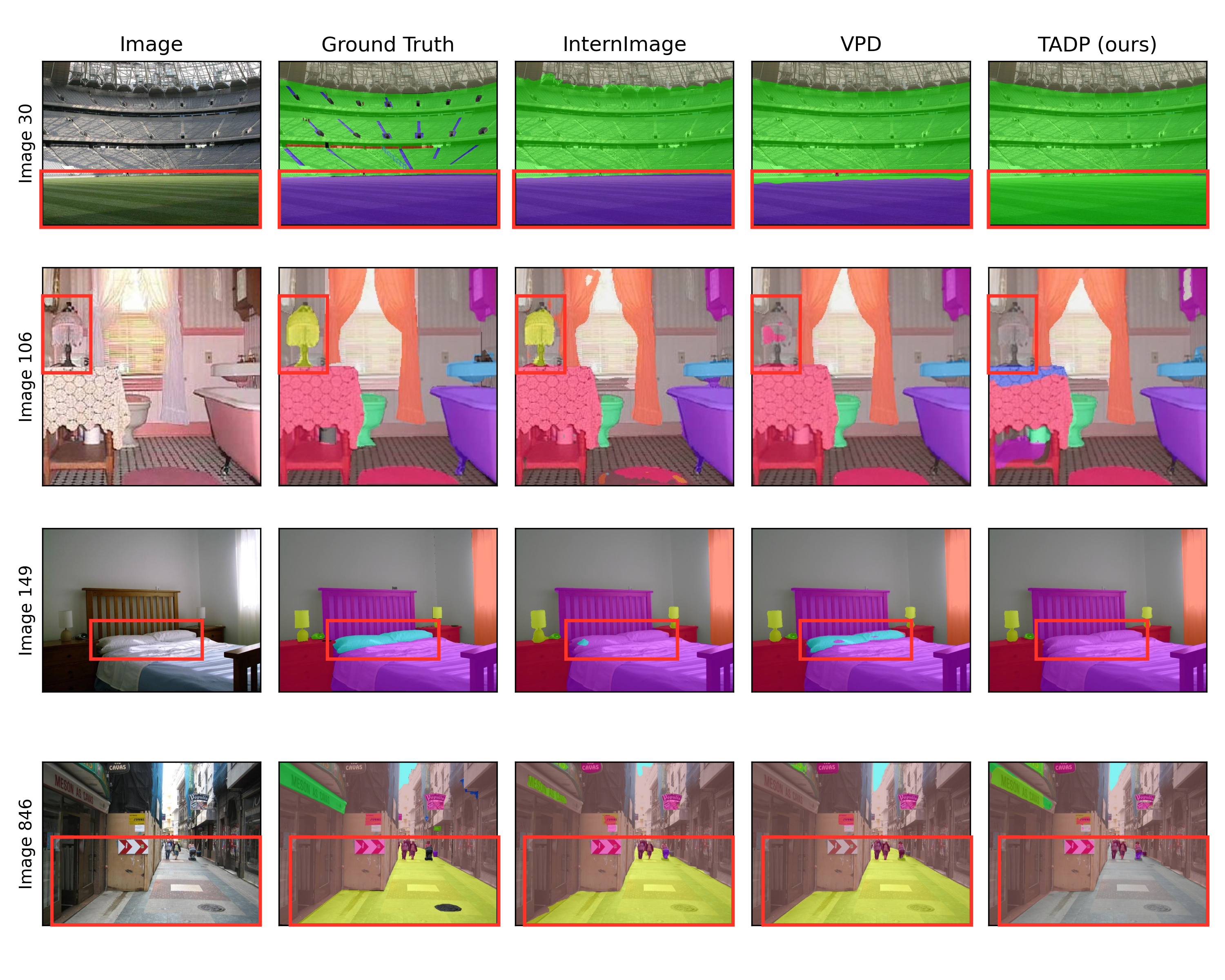}
    \caption{\textbf{Image Segmentation Comparison: Image, Ground Truth, and Prediction visualizations for InternImage, VPD, and TADP (ours) in ADE20K.}
    Red boxes show where TADP is worse than InternImage and/or VPD.}
\end{figure*}

\begin{figure*}[h]
    \centering 
    \includegraphics[width=1\textwidth]{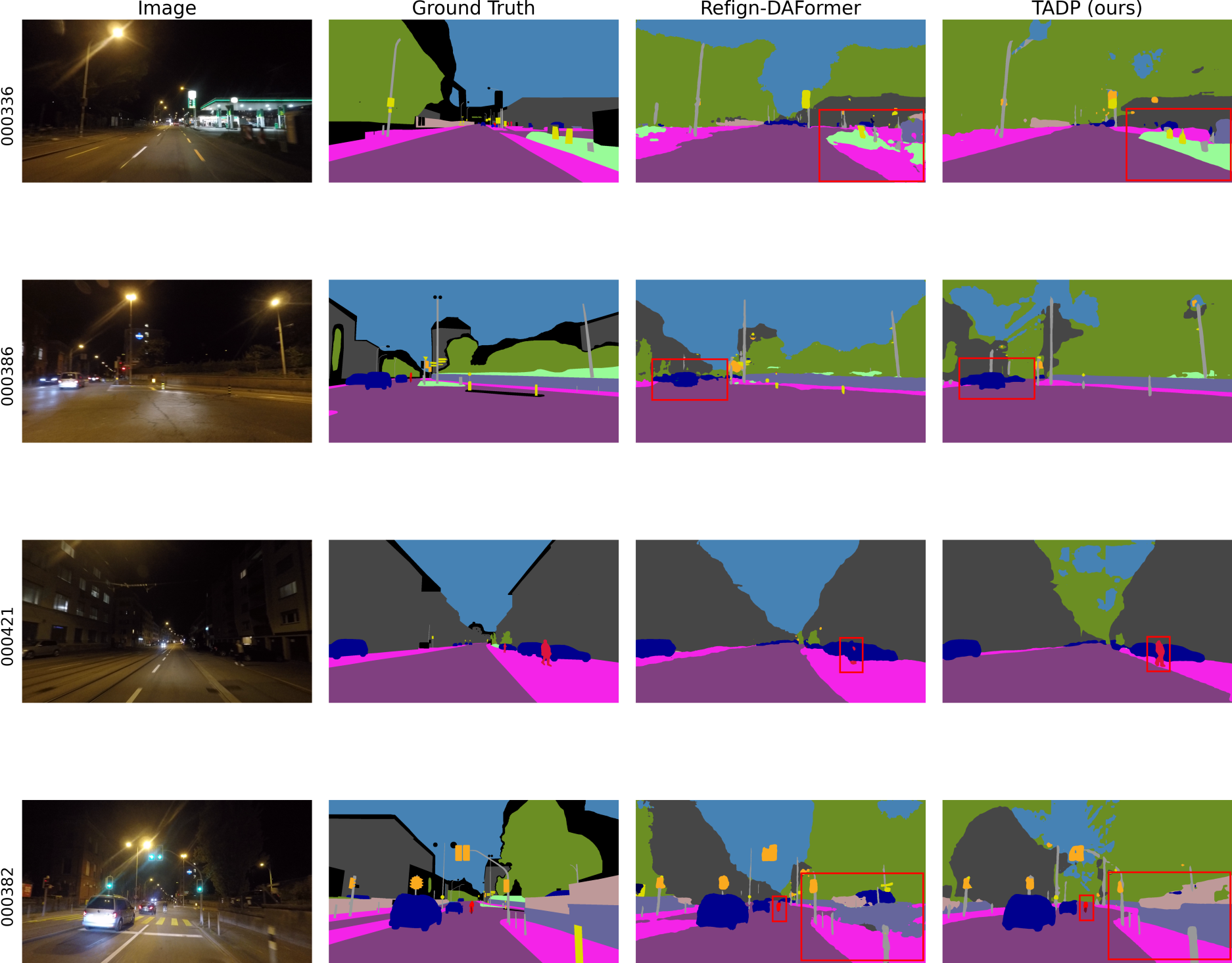}
    \caption{\textbf{Cross-domain Image Segmentation Comparison: Image, Ground Truth, and Prediction visualizations for Refign-DAFormer, and TADP (ours) for Cityscapes to Dark Zurich Val.}
    Red boxes show where TADP is better than Refign-DAFormer.}
\end{figure*}

\begin{figure*}[h]
    \centering 
    \includegraphics[width=1\textwidth]{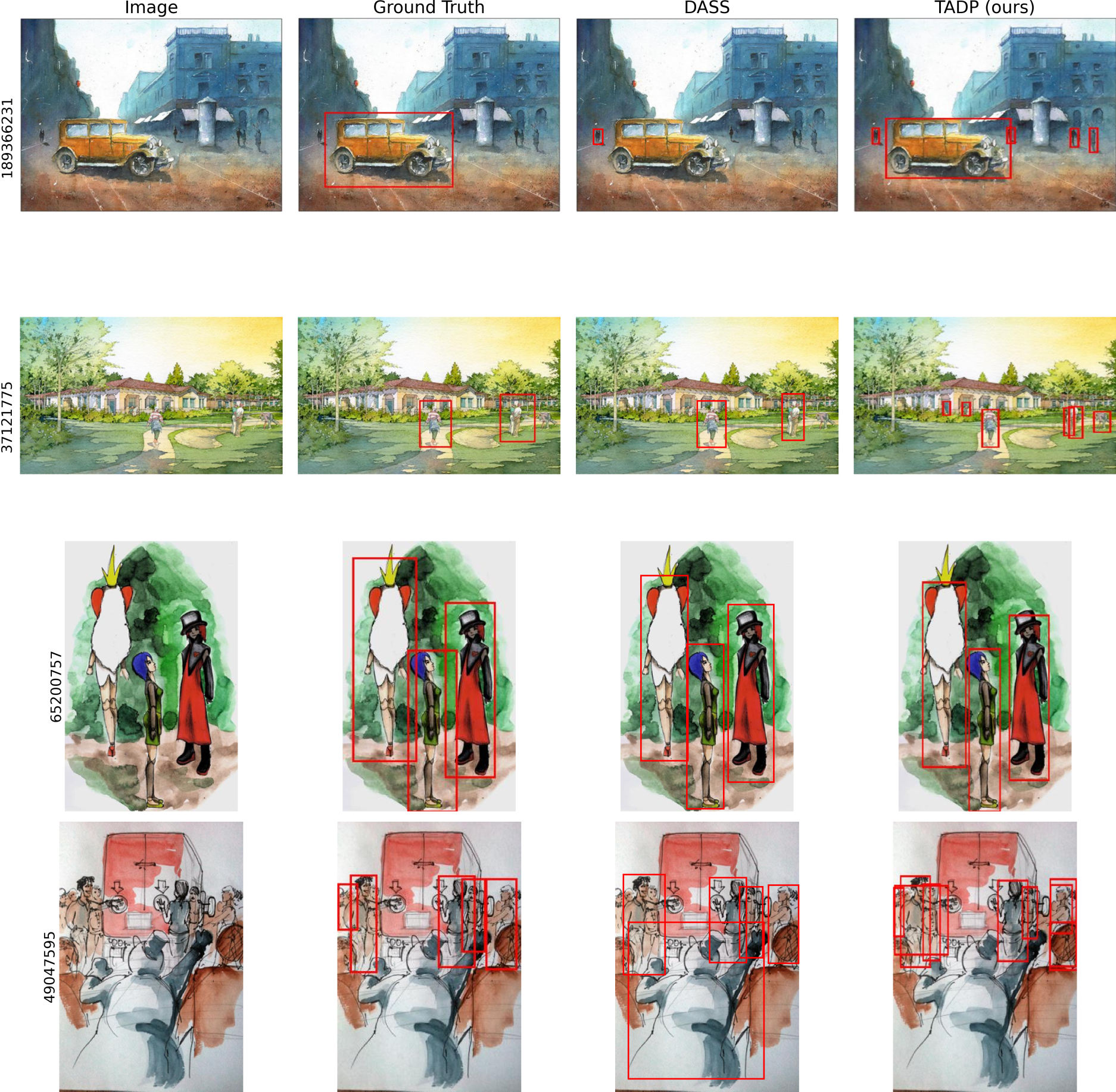}
    \caption{\textbf{Cross-domain Object Detection Comparison: Image, Ground Truth, and Prediction visualizations for DASS, and TADP (ours) for Pascal VOC to Watercolor2k.}
    Red boxes show the detections of each model. Notice that TADP not only beats DASS mostly, but also finds more objects than the ones annotated in the ground truth. }
\end{figure*}

\newpage
\clearpage

\section{Implementation Details}
\label{appendix:implementation_details}

To isolate the effects of our text-image alignment method, we ensure our model setup precisely follows prior work. Following VPD \citep{zhao_unleashing_2023}, we jointly train the task-specific head and the diffusion backbone. The learning rate of the backbone is set to 1/10 the learning rate of the head to preserve the benefits of pre-training better. We describe the different tasks by describing $H$ and $\mathcal{L}_H$. We use an FPN \citep{kirillov_panoptic_2019} head with a cross-entropy loss for segmentation. We use the same convolutional head used in VPD for monocular depth estimation with a Scale-Invariant loss \citep{eigen_depth_2014}. For object detection, we use a Faster-RCNN head with the standard Faster-RCNN loss \citep{ren_faster_2017}\footnote{Object detection was not explored in VPD.}. Further details of the training setup can be found in \cref{tab:hyperparams-single} and \cref{tab:hyperparams-cross}. In our single-domain tables, we include our reproduction of VPD, denoted with a (R). We compute our relative gains with our reproduced numbers, with the same seed for all experiments.

\vspace{2cm} 

\input{tables/hyperparameters_subtables}

\FloatBarrier

\subsection{Model personalization}
\label{appendix:model_personalization}
For textual inversion, we use 500 images from DZ-train and five images for W2K and C2K and train all tokens for 1000 steps. We use a constant learning rate scheduler with a learning rate of $5e-4$ and no warmup. For Dreambooth, we use the same images as in textual inversion but train the model for 500 steps (DZ) steps or 1000 steps (W2K and C2K). We use a learning rate of $2e-6$ with a constant learning rate scheduler and no warmup. We use no prior preservation loss.

\clearpage

\end{document}

%% file: preamble.tex
\usepackage[dvipsnames]{xcolor}


%% file: tables/seg_pascal.tex
\begin{table}[t]
  \centering

    \small
    \begin{tabularx}{\linewidth}{lCCCCCC}
    \toprule
    Method & Avg & TA & LS & G & OT & mIoU\textsuperscript{ss} \\
    \midrule
    VPD(R) \citep{zhao_unleashing_2023} & \checkmark & \checkmark & & & \checkmark & 82.34 \\
    VPD(LS) & \checkmark & \checkmark & \checkmark & & \checkmark & 83.06 \\
    Class Embs & & & \checkmark &  & \checkmark & 82.72 \\
    Class Names & & & \checkmark & & \checkmark & 84.08 \\
    \ours-0  & & & \checkmark & \checkmark & & 86.36 \\
    \ours-20  & & & \checkmark & \checkmark & & 86.19 \\
    \ours-40  & & & \checkmark & \checkmark & & \textbf{87.11} \\
    \ours(NO)-20  & & & \checkmark & & & 86.35 \\
    \midrule
    \textit{\ours-Oracle} & & & \checkmark & & & 89.85 \\
    \bottomrule
    \end{tabularx}
\vspace{-.2cm}
    \caption{\textbf{Prompting for Pascal VOC2012 Segmentation.} We report the single-scale validation mIoU for Pascal experiments. (R): Reproduction of VPD, Avg: EOS token averaging, LS: Latent Scaling, G: Grammar, OT: Off-target information. For our method, we indicate the minimum length of the BLIP caption with \ours-$X$ and nouns only with (NO).}
  \label{tab:pascal}%

\end{table}%

%% file: tables/seg_ade20k_full.tex

\definecolor{LightGreen}{RGB}{237,249,235}

\begin{table}[t]
  \centering
  \adjustbox{width=1\linewidth}{
    \begin{tabular}{lccccc}
    \toprule
    Method & \#Params & FLOPs & Crop & mIoU$^{\rm ss}$ & mIoU$^{\rm ms}$ \\
    \midrule
    \addlinespace[3.0pt]
    \multicolumn{6}{l}{\textit{self-supervised pre-training}}\\
    EVA \citep{fang_eva_2022} & 1.01B & - & 896$^2$ & 61.2 & 61.5 \\
    InternImage-L \citep{wang_internimage_2022} & 256M & 2526G & 640$^2$ & 53.9 & 54.1 \\ 
    InternImage-H \citep{wang_internimage_2022} & 1.31B & 4635G & 896$^2$ & \cellcolor{CBLightYellow}\textbf{62.5} & \textbf{62.9} \\ 
    \midrule
    \addlinespace[3.0pt]
    \multicolumn{6}{l}{\textit{multi-modal pre-training}}\\
    CLIP-ViT-B \citep{rao_denseclip_2022} & 105M & 1043G & 640$^2$ &  50.6 & 51.3 \\
    ViT-Adapter \citep{chen_vision_2022} & 571M & - & 896$^2$ & 61.2 & 61.5 \\
    BEiT-3 \citep{wang_image_2023} & 1.01B & - & 896$^2$ & \textbf{62.0} & 62.8 \\
    ONE-PEACE \citep{wang_one-peace_2023} & 1.52B & - & 896$^2$ & \textbf{62.0} & \cellcolor{CBLightYellow}\textbf{63.0} \\    
    \midrule
    \addlinespace[3.0pt]
    \multicolumn{6}{l}{\textit{diffusion-based pre-training}}\\
    VPD$_{\rm A32}$ \citep{zhao_unleashing_2023}   & 862M & 891G & 512$^2$ & 53.7  & 54.6 \\
    VPD(R)   & 862M & 891G & 512$^2$ & 53.1  & 54.2 \\
    VPD(LS)   & 862M & 891G & 512$^2$ & 53.7  & 54.4 \\
    \rowcolor{CBLightGreen} \ours-40 (Ours) & 862M & 2168G & 512$^2$ & \textbf{54.8}  &  \textbf{55.9} \\
    \midrule
    \textit{\ours-Oracle} & 862M & - & 512$^2$ & 72.0  & - \\
    \bottomrule
    \end{tabular}%
  } 
  \caption{\textbf{Semantic segmentation with different methods for ADE20k.} Our method (green) achieves SOTA within the diffusion-pretrained models category. The results of our oracle indicate the potential of diffusion-based models for future research as it is significantly higher than the overall SOTA (highlighted in yellow). See \cref{tab:pascal} for a notation key and \cref{tab:seg_ade_fast} for fast schedule results.
  }
  \label{tab:ade}%
\end{table}%


%% file: tables/depth_nyu.tex
\begin{table}
  \centering

  \adjustbox{width=1\linewidth}{
    \begin{tabular}{lcccccc}
    \toprule
    Method & RMSE$\downarrow$  & $\delta_1\uparrow$    & $\delta_2\uparrow$    & $\delta_3\uparrow$    & REL $\downarrow$  & log10 $\downarrow$ \\\midrule
    \multicolumn{7}{l}{\textit{default schedule}}\\
    \midrule
    SwinV2-L \citep{liu_swin_2021-1} & 0.287 & 0.949 & 0.994 & 0.999 & 0.083 & 0.035 \\
    AiT \citep{ning_all_2023} & 0.275 & 0.954 & 0.994 & 0.999 & 0.076 & 0.033\\
    ZoeDepth \citep{bhat_zoedepth_2023} & 0.270 & 0.955 & 0.995 & 0.999 & 0.075 & 0.032\\
    VPD \citep{zhao_unleashing_2023} & 0.254 & 0.964 & 0.995 & 0.999 & 0.069 & 0.030 \\
    \midrule
    VPD(R) & 0.248 & 0.965 & 0.995 & 0.999 & 0.068  & 0.029 \\
    VPD(LS) & 0.235 & 0.971 & 0.996 & 0.999 & 0.064  & 0.028 \\
    \rowcolor{CBLightGreen}\ours-40 & \textbf{0.225} & \textbf{0.976} & \textbf{0.997} & \textbf{0.999} & \textbf{0.062} & \textbf{0.027} \\
    \bottomrule
    \addlinespace[4.0pt]
    \multicolumn{7}{l}{\textit{fast schedule, 1 epoch}}\\
    \midrule
    VPD & 0.349 & 0.909 & 0.989 & 0.998 & 0.098 & 0.043 \\
    VPD(R) & 0.340 & 0.910 & 0.987 & 0.997 & 0.100 & 0.042\\
    VPD(LS) & 0.332 & 0.926 & 0.992 & 0.998 & 0.097 & 0.041 \\
    \rowcolor{CBLightGreen}\ours-0 & \textbf{0.328} & \textbf{0.935} & \textbf{0.993} & \textbf{0.999} & \textbf{0.082} & \textbf{0.038} \\
    \bottomrule
    \end{tabular}
    }
    \caption{\textbf{Depth estimation in NYUv2.} We find latent scaling accounts for a relative gain of $\sim5.5\%$ on the RMSE metric. Additionally, image-text alignment improves  $\sim4\%$ relative on the RMSE metric. A minimum caption length of 40 tokens performs the best.We also explore adding a text-adapter (TA) to TADP, but find no significant gain. See Table~\ref{tab:pascal} for a notation key.}
    \label{tab:depth}
\end{table}

%% file: tables/crossdomain_cityscapes.tex
\begin{table}[t]
  \centering

  \begin{tabularx}{\linewidth}{lCCCCCC}
  \toprule
    \multirow{2}{*}{Method} & Dark Zurich-val & ND \\
    &  mIoU &  mIoU \\
    \midrule
    DAFormer \citep{hoyer_daformer_2022} & -- & 54.1 \\
    Refign-DAFormer \citep{bruggemann_refign_2022} & -- & 56.8 \\
    PTDiffSeg \citep{gong_prompting_2023} & 37.0 & -- \\
    \midrule
    \midrule
    \ours\textsubscript{null} & \textbf{42.8} & 57.5 \\
    \ours\textsubscript{simple} & 39.1 & 56.9 \\
    \ours\textsubscript{TextualInversion} & 41.4 & \textbf{60.8} \\
    \ours\textsubscript{DreamBooth} & 38.9 & 60.4 \\
    \midrule 
    \ours\textsubscript{NearbyDomain} & 41.9 & 56.9 \\
    \ours\textsubscript{UnrelatedDomain} & 42.3 & 55.1 \\
    \bottomrule
  \end{tabularx}
  \caption{\textbf{Cross-domain semantic segmentation.} Cityscapes (CD) to Dark Zurich (DZ) val and Nighttime Driving (ND). We report the mIoU. Our method sets a new SOTA for DarkZurich and Nighttime Driving.
  }
  \label{tab:cross-domain-cityscapes}
\end{table}

%% file: tables/crossdomain_pascal.tex
\begin{table}[t]
  \centering
  \small

  \adjustbox{width=1.0\linewidth}{
  \vspace{3pt}
  \begin{tabular}{lcccccc}
  \toprule
    \multirow{2}{*}{Method} & \multicolumn{2}{c}{Watercolor2k} & \multicolumn{2}{c}{Comic2k} \\
    & AP & AP$_{50}$ & AP & AP$_{50}$ \\
    \midrule
    \multicolumn{5}{l}{\textit{Single Domain Generalization (SGD)}}\\
    \midrule
    CLIP the gap \citep{vidit_clip_2023}   & -- & 33.5 & -- & 43.4 \\
    \midrule
    \multicolumn{5}{l}{\textit{Cross domain weakly supervised object detection}}\\
    \midrule
    PLGE \citep{ouyang2021pseudo}   & -- & 56.5 & -- & 41.7 \\
    ICCM \citep{hou2021informative}   & -- & 57.4 & -- & 37.1 \\
    H2FA R-CNN \citep{xu2022h2fa}  & -- & 59.9 & -- & 46.4 \\
    \midrule
    \multicolumn{5}{l}{\textit{Unsupervised domain adaptation object detection}}\\
    \midrule
    ADDA \citep{tzeng2017adversarial}  & -- & 49.8 & -- &  23.8 \\
    MCAR \citep{zhao2020adaptive}  & -- & 56.0 & -- & 33.5 \\
    UMT \citep{deng2021unbiased}  & -- & 58.1 & -- & -- \\
    DASS-Detector (extra data) \citep{topal2022domain}  & -- & 71.5 & -- & \cellcolor{CBLightYellow}\textbf{64.2} \\
    \midrule
    \midrule
    \ours\textsubscript{null} & 42.1 & 72.1 & 31.1 & \textbf{57.4} \\
    \ours\textsubscript{simple} & \textbf{43.5} & \textbf{72.2} & 31.9 & 56.6\\
    \ours\textsubscript{TextualInversion} & 43.2 & \textbf{72.2} & \textbf{33.2} & \textbf{57.4}\\
    \ours\textsubscript{DreamBooth} & 43.2 & \textbf{72.2} & 32.9 &  56.9\\
    \midrule 
    \ours\textsubscript{NearbyDomain} & 42.0 & 71.5 & 31.8 & 56.4\\
    \ours\textsubscript{UnrelatedDomain} & 42.2 & 71.9 & 32.0 & 55.9\\
    \bottomrule
  \end{tabular}
  }
  \caption{\textbf{Cross-domain object detection.} Pascal VOC to Watercolor2k and Comic2k. We report the AP and $AP_{50}$. Our method sets a new SOTA for  Watercolor2K.
  }
  \label{tab:cross-domain-pascal}
\end{table}

%% file: tables/oracle_ablations.tex
\begin{table}[ht!]
    \centering

    \small
    \adjustbox{width=0.3\linewidth}{
    \begin{tabular}{cc}
        & \textbf{Recall} \\
        \rotatebox[origin=c]{90}{\textbf{Precision}} &
        \begin{tabular}{c|c|c|c}
            & \textbf{0.50} & \textbf{0.75} & \textbf{1.00} \\
            \hline
            \rotatebox[origin=c]{90}{\textbf{  0.50  }} & 49.53 & 52.00 & 55.22 \\
            \hline
            \rotatebox[origin=c]{90}{\textbf{  0.75  }} & 49.17 & 51.46 & 58.62 \\
            \hline
            \rotatebox[origin=c]{90}{\textbf{  1.00  }} & 50.20 & 54.82 & \cellcolor{CBLightCyan} 63.29 \\
        \end{tabular} \\
    \end{tabular}
    }
    \caption{\textbf{ADE20K - Oracle Precision-Recall Ablations} We modify the oracle captions by randomly adding or removing classes such that the precision and recall are 0.50, 0.75, or 1.00. We train models on ADE20K on a fast schedule (4K) using these captions. The 4k iteration oracle equivalent is highlighted in blue.}
    \label{tab:oracle_ablations}
\end{table}

%% file: tables/hyperparameters_subtables.tex
\begin{table}[ht!]
\begin{multicols}{2}
\begin{subtable}{\linewidth}
\centering
\vspace{12pt}
\begin{tabularx}{\linewidth}{l|X}
    \textbf{Hyperparameter} & \textbf{Value} \\
\cline{1-2}
    Learning Rate & $0.00008$ \\
    Batch Size & $2$ \\
    Optimizer & AdamW \\
    Weight Decay & $0.005$ \\
    Warmup Iters & $1500$ \\
    Warmup Ratio & $1e-6$ \\
    U-Net Learning Rate Scale & $0.01$ \\
    Training Steps & $80000$ \\
\end{tabularx}
\caption{ADE20k - full schedule}
\label{tab:hyperparameters_ade20k_full}
\end{subtable}

\begin{subtable}{\linewidth}
\centering
\vspace{12pt}
\begin{tabularx}{\linewidth}{l|X}
    \textbf{Hyperparameter} & \textbf{Value} \\
\cline{1-2}
    Learning Rate & $0.00016$ \\
    Batch Size & $2$ \\
    Optimizer & AdamW \\
    Weight Decay & $0.005$ \\
    Warmup Iters & $150$ \\
    Warmup Ratio & $1e-6$ \\
    Unet Learning Rate Scale & $0.01$ \\
    Training Steps & $8000$ \\
\end{tabularx}
\caption{ADE20k - fast schedule 8k}
\label{tab:hyperparameters_ade20k_fast8k}
\end{subtable}

\begin{subtable}{\linewidth}
\centering
\vspace{12pt}
\begin{tabularx}{\linewidth}{l|X}
    \textbf{Hyperparameter} & \textbf{Value} \\
\cline{1-2}
    Learning Rate & $0.00016$ \\
    Batch Size & $2$ \\
    Optimizer & AdamW \\
    Weight Decay & $0.005$ \\
    Warmup Iters & $75$ \\
    Warmup Ratio & $1e-6$ \\
    Unet Learning Rate Scale & $0.01$ \\
    Training Steps & $4000$ \\
\end{tabularx}
\caption{ADE20k - fast schedule 4k}
\label{tab:hyperparameters_ade20k_fast4k}
\end{subtable}

\begin{subtable}{\linewidth}
\centering
\vspace{12pt}
\begin{tabularx}{\linewidth}{l|X}
    \textbf{Hyperparameter} & \textbf{Value} \\
\cline{1-2}
    Learning Rate & $5e-4$ \\
    Batch Size & $3$ \\
    Optimizer & AdamW \\
    Weight Decay & $0.1$ \\
    Layer Decay & $0.9$ \\
    Epochs & 25 \\
    Drop Path Rate & $0.9$ \\
\end{tabularx}
\caption{NYUv2}
\label{tab:hyperparameters_nyuv2}
\end{subtable}

\begin{subtable}{\linewidth}
\centering
\vspace{12pt}
\begin{tabularx}{\linewidth}{l|X}
    \textbf{Hyperparameter} & \textbf{Value} \\
\cline{1-2}
    Learning Rate & $5e-4$ \\
    Batch Size & $3$ \\
    Optimizer & AdamW \\
    Weight Decay & $0.1$ \\
    Layer Decay & $0.9$ \\
    Epochs & 1 \\
    Drop Path Rate & $0.9$ \\
\end{tabularx}
\caption{NYUv2 - fast schedule }
\label{tab:hyperparameters_nyuv2_fast}
\end{subtable}

\begin{subtable}{\linewidth}
\centering
\vspace{12pt}
\begin{tabularx}{\linewidth}{l|X}
    \textbf{Hyperparameter} & \textbf{Value} \\
\cline{1-2}
    Learning Rate & $0.00001$ \\
    Batch Size & $2$ \\
    Gradient Accumulation & $4$ \\
    Epochs & $15$ \\
    Optimizer & AdamW \\
    Weight Decay & $0.01$ \\
\end{tabularx}
\caption{Pascal VOC}
\label{tab:hyperparameters_pascal_voc}
\end{subtable}
\end{multicols}
\caption{\textbf{Single-Domain Hyperparameters.}}
\label{tab:hyperparams-single}
\end{table}

\begin{table}[ht!]

\begin{multicols}{2}
\begin{subtable}{\linewidth}
\centering
\vspace{12pt}
\begin{tabularx}{\linewidth}{l|X}
    \textbf{Hyperparameter} & \textbf{Value} \\
\cline{1-2}
    Learning Rate & $0.00008$ \\
    Batch Size & $2$ \\
    Optimizer & AdamW \\
    Weight Decay & $0.005$ \\
    Warmup Iters & $1500$ \\
    Warmup Ratio & $1e-6$ \\
    Unet Learning Rate Scale & $0.01$ \\
    Training Steps & $40000$ \\
\end{tabularx}
\caption{Cityscapes $\rightarrow$ Dark Zurich \& NightTime Driving}
\label{tab:hyperparameters_cityscapes}
\end{subtable}

\begin{subtable}{\linewidth}
\centering
\vspace{12pt}
\begin{tabularx}{\linewidth}{l|X}
    \textbf{Hyperparameter} & \textbf{Value} \\
\cline{1-2}
    Learning Rate & $0.00001$ \\
    Batch Size & $2$ \\
    Epochs & $100$ \\
    Optimizer & AdamW \\
    Weight Decay & $0.01$ \\
    Learning Rate Schedule & Lambda \\
\end{tabularx}
\caption{Pascal VOC $\rightarrow$ Watercolor \& Comic}
\label{tab:hyperparameters_pascal_voc_wc}
\end{subtable}

\begin{subtable}{\linewidth}
\centering
\vspace{12pt}
\begin{tabularx}{\linewidth}{l|X}
    \textbf{Hyperparameter} & \textbf{Value} \\
\cline{1-2}
    Prior Preservation Cls Images & $200$ \\
    Learning Rate & $5e-6$ \\
    Training Steps & $1000$ \\
\end{tabularx}
\caption{Dreambooth Hyperparameters}
\label{tab:hyperparameters_dreambooth}
\end{subtable}

\begin{subtable}{\linewidth}
\centering
\vspace{12pt}
\begin{tabularx}{\linewidth}{l|X}
    \textbf{Hyperparameter} & \textbf{Value} \\
\cline{1-2}
    Steps & $3000$ \\
    Learning Rate &  $5.0e-04$ \\
    Batch Size & $1$ \\
    Gradient Accumulation & $4$ \\
\end{tabularx}
\caption{Textual Inversion Hyperparameters}
\label{tab:hyperparameters_textual}
\end{subtable}
\end{multicols}
\caption{\textbf{Cross-Domain Hyperparameters.}}
\label{tab:hyperparams-cross}
\end{table}